\documentclass[letterpaper]{article} 
\usepackage[preprint]{aaai2027}  
\usepackage[hyphens]{url}  
\usepackage{graphicx} 
\usepackage{natbib}  
\usepackage{caption} 
\usepackage{algorithm}
\usepackage{algorithmic}
\usepackage{multirow}
\usepackage{makecell}
\usepackage{pifont}
\usepackage{subcaption}
\usepackage{amsmath}
\usepackage{amsfonts}
\usepackage[table]{xcolor}
\definecolor{lightCyan}{rgb}{0.95,0.95,0.95}

\usepackage{newfloat}
\usepackage{listings}
\DeclareCaptionStyle{ruled}{labelfont=normalfont,labelsep=colon,strut=off} 
\floatstyle{ruled}
\newfloat{listing}{tb}{lst}{}
\floatname{listing}{Listing}

\usepackage{booktabs}

\title{Breaking the Horizontal Prior: From Long-Tailed Orientation Bias to \\ Roll-Robust Monocular Depth Estimation}

\author{
    Kaihua Tang$^1$ \quad Ziqing Xia$^1$\corresponding \quad Xiaoxu Zheng$^2$ \quad Xiaoxue Zhang$^2$ \quad Michael Bi Mi$^2$\\
    Zhan Xu$^2$ \quad Dave Zhenyu Chen$^2$ 
}
\affiliations{
    \textsuperscript{\rm 1}Tongji University \quad \textsuperscript{\rm 2}Huawei Technologies Ltd.\\

}

\begin{document}

\maketitle

\begin{abstract}

Despite recent advances in Monocular Depth Estimation, state-of-the-art depth foundation models remain vulnerable to robustness issues. Particularly, even slight camera rolls can result in substantial degradation in depth estimations. We attribute this problem to a previously overlooked phenomenon, termed the \textbf{Horizontal Prior}, which is a manifestation of long-tailed distribution bias: most training images are captured in approximately horizontal orientations due to human visual preferences and photographic habits. While intuitive remedies such as re-balanced data augmentation and horizon leveling provide partial improvements, they fail to fully address the issue. In this paper, we introduce Invariant Depth Constraint (ID-Constraint), a training-time supervision strategy that improves roll robustness by fine-tuning and jointly regularizing the depth backbone with a series of geometric and spatial reasoning tasks. These auxiliary objectives encourage the backbone to learn rotation-stable, depth-relevant representations, while the auxiliary prediction heads are discarded after training, leaving the original inference architecture unchanged. Extensive experiments on five benchmark datasets across four roll settings demonstrate the effectiveness of the proposed method.~\footnote{Code Link: https://github.com/KaihuaTang/Horizontal-Prior}

\end{abstract}

\section{Introduction}
\label{sec:intro}

\begin{figure}[!t]
    \centering
    \includegraphics[width=1.0\linewidth]{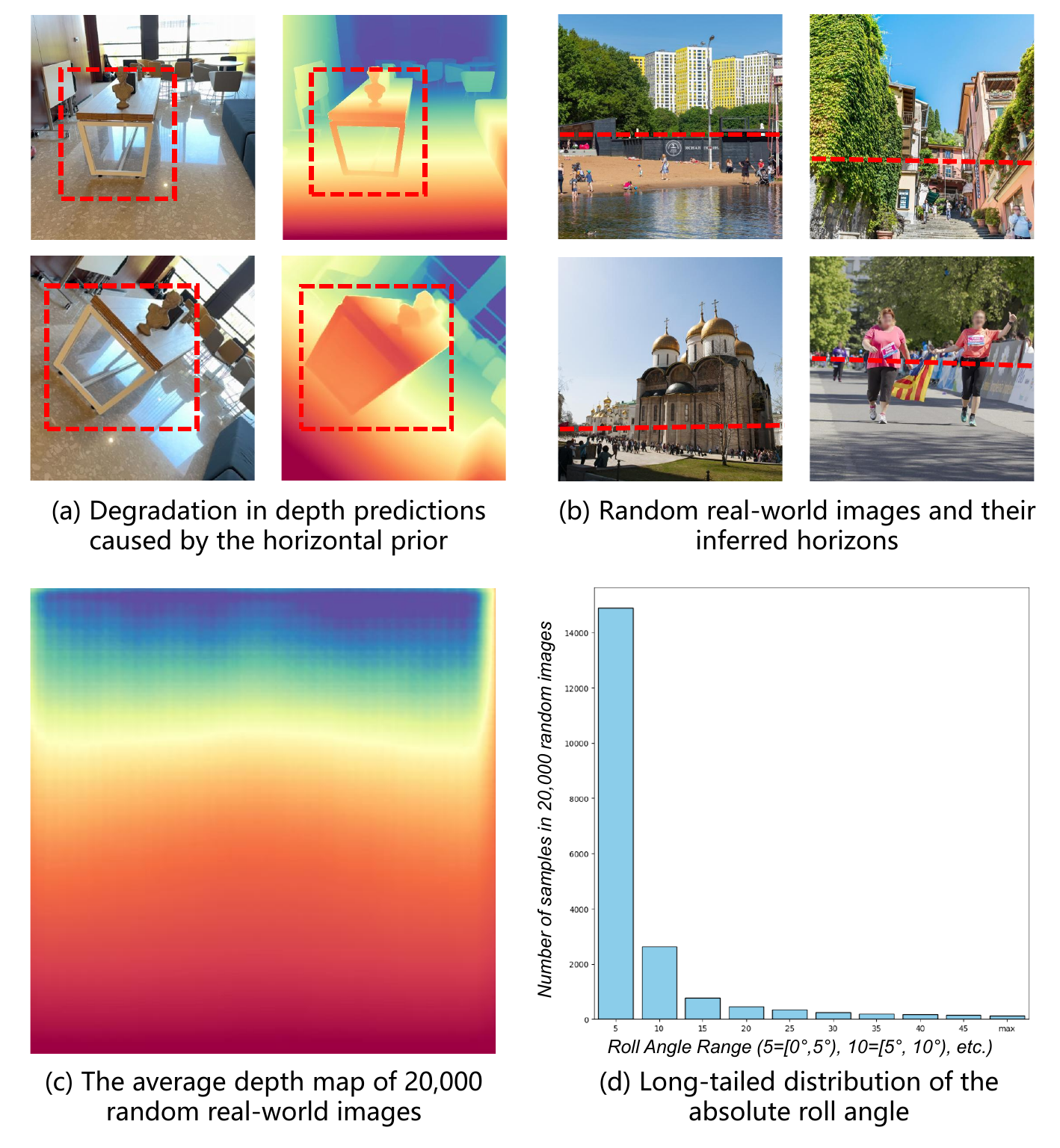}
    \caption{Investigating the Horizontal Prior: (a) images with non-horizontal orientations exhibit substantial degradation in depth predictions; (b) randomly sampled images demonstrate that the visual data are mostly horizontal; (c) the average depth map and (d) the long-tailed distribution of absolute roll angle of 20K random real-world images further provides qualitative and quantitative evidence of the Horizontal Prior.}
    \label{fig1:longtail}
\end{figure}

\begin{figure}[tb]
    \centering
    \includegraphics[width=1.0\linewidth]{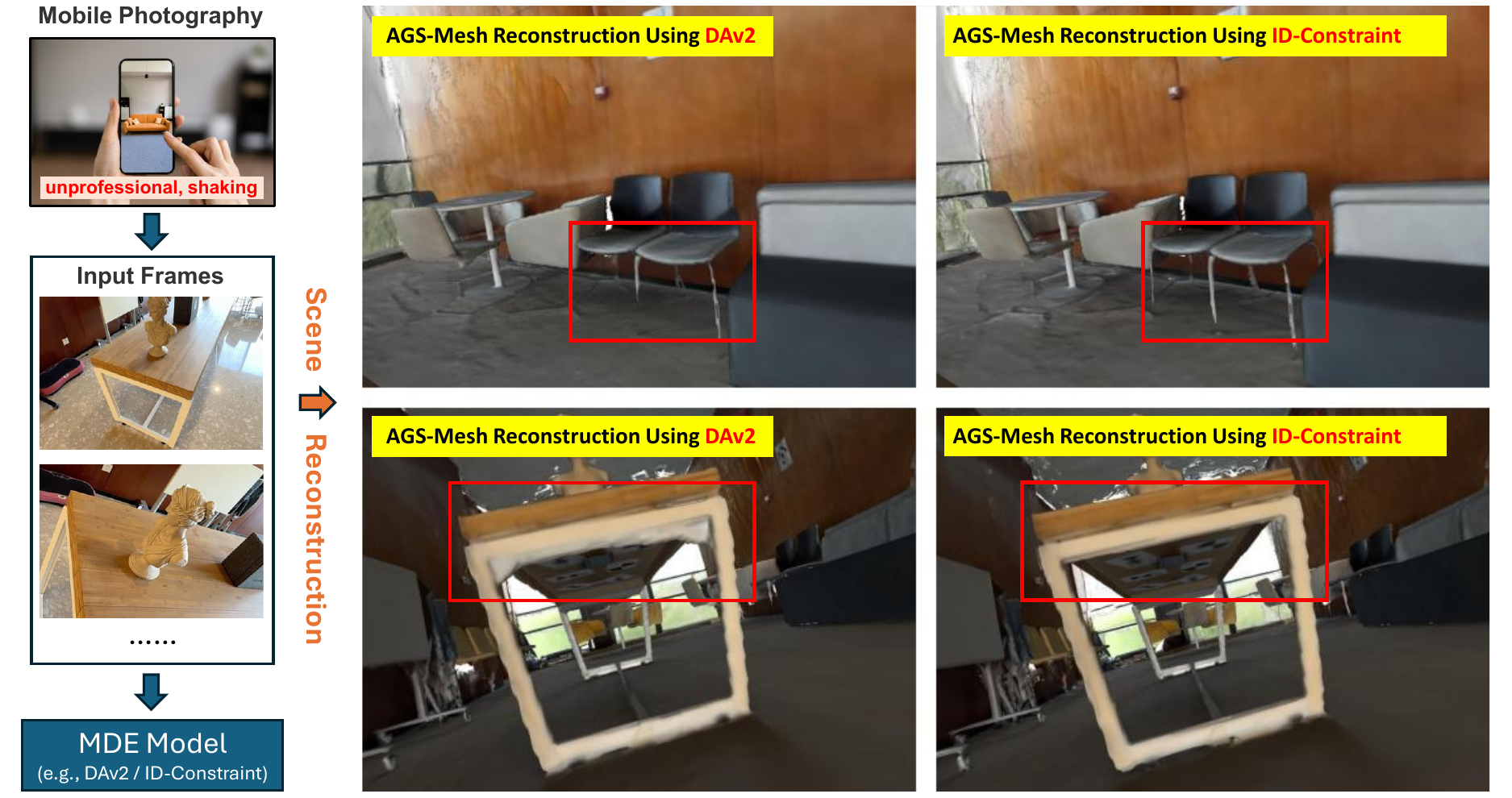}
    \caption{3D reconstruction flaws caused by the horizontal prior in real-world applications.}
    \label{fig2:case}
\end{figure}

Monocular Depth Estimation (MDE)~\cite{ming2021deep,arampatzakis2023monocular} is a fundamental problem in computer vision with broad downstream applications~\cite{yan2024monocd,guo2025multi,liu2025multi,hurtado2025panoptic}. Recent advances in visual foundation models~\cite{radford2021clip,dosovitskiy2020image,oquab2024dinov2,zhai2023sigmoid} and generative diffusion models~\cite{croitoru2023diffusion,xu2025genpercept,ke2023marigold} have led to notable improvements in MDE performance. However, in real-world scenarios where images are often casually captured with mobile devices, these models tend to exhibit significant robustness degradation. In particular, even slight camera shake can cause substantial instability in the estimated depth maps, as illustrated in Figure~\ref{fig1:longtail}(a), leading to noticeable artifacts in downstream applications, as shown in Figure~\ref{fig2:case}. This sensitivity to minor perturbations severely undermines the practical applicability of state-of-the-art MDE models.

In this paper, we attribute the observed robustness issue in Figure~\ref{fig2:case} to the \textbf{Horizontal Prior}, which is a manifestation of the long-tailed distribution bias in MDE. As shown in Figure~\ref{fig1:longtail}(b), we randomly sample several real-world images from the large-scale SA-1B dataset~\cite{kirillov2023segany} and observe that their inferred horizons are all approximately horizontal. To obtain statistically meaningful evidence, we further sample 20,000 random images and estimate their depth maps together with their absolute roll magnitudes. The average depth map and long-tailed roll distribution shown in Figure~\ref{fig1:longtail}(c,d) further support the existence of the horizontal prior. This bias arises from professional photography conventions and inherent human perceptual preferences~\cite{hansen2004horizontal,luo2003psychophysical}. However, in practice, perfect horizontal alignment is rarely achieved. Handheld camera shake, intentional motion, and cinematic transitions inevitably lead to a long-tailed distribution of roll magnitudes. Consequently, MDE models trained on such biased data exhibit substantial performance degradation when encountering underrepresented non-horizontal inputs.

To address this robustness issue induced by the horizontal prior, two intuitive approaches can be considered: re-balanced data augmentation and horizon leveling. These correspond respectively to re-balanced training~\cite{he2021re,tang2022invariant,kang2019decoupling} and training-free adjustment~\cite{menon2020long,tang2020long} methods in long-tailed classification~\cite{zhang2023deep}. Specifically, the former attempts to re-balance the training distribution of image roll angles through data augmentation. Although this approach can partially alleviate robustness issues caused by the horizontal prior, it also introduces new challenges for feature learning, as reported in prior long-tailed learning studies~\cite{kang2019decoupling,tang2020long}. In particular, directly applying re-balancing to MDE training data may exacerbate the domain gap between DINOv2 pretraining~\cite{oquab2024dinov2} and MDE fine-tuning distributions. 
Horizon leveling represents another potential remedy, which seeks to rotate images to achieve horizontal alignment. While conceptually straightforward, this process is far from trivial. Commercial solutions~\cite{GoPro_HorizonLeveling_2022} typically rely on auxiliary hardware sensors (\textit{e.g.}, six-axis IMU) for orientation estimation~\cite{AhmadGhazilla2013_IMUReview}. In contrast, purely algorithmic methods~\cite{WorkmanZhaiJacobs2016_HLW} suffer from substantial alignment errors, ranging from 25.90$^{\circ}$ to 47.94$^{\circ}$ on average as reported in Table~\ref{tab4:angle}, which already leads to significant degradation in MDE performance.

To address these limitations, we propose a novel algorithm that enhances the robustness of MDE models by leveraging rotation-invariant auxilliary tasks. Recent studies~\cite{danier2025depthcues} have shown that the performance of existing MDE models is positively correlated with the capability of their vision backbones to capture depth cues, which are assessed through a series of geometric and spatial reasoning tasks. We notice that a subset of these tasks are inherently rotation-invariant, suggesting that foundation vision models such as DINOv2~\cite{oquab2024dinov2} have implicitly learned depth-related invariant representations. Building on this insight, we hypothesize that encouraging the model to rely more on such invariant features during depth prediction can naturally mitigate the robustness issues induced by the horizontal prior. However, the existing benchmark~\cite{danier2025depthcues} primarily focuses on region-level predictions and lacks fine-grained, pixel-level supervision. To bridge this gap, we introduce two additional pixel-level supervision tasks to reinforce feature learning at finer resolutions. We therefore propose the Invariant Depth Constraint (ID-Constraint), which employs auxiliary heads and losses to promote rotation-invariant feature learning. These auxiliary heads are removed during inference, introducing no additional computational overhead.

In our experiments, we systematically investigate the robustness degradation caused by the horizontal prior and evaluate the effectiveness of the proposed algorithms across five benchmark datasets: DIODE~\cite{vasiljevic2019diode}, ScanNet~\cite{dai2017scannet}, ETH3D~\cite{schops2017multi}, KITTI~\cite{geiger2012we}, and NYUv2~\cite{silberman2012indoor}. To facilitate detailed analysis, we define four test settings based on absolute roll angles: Horizontal ($0^{\circ}$), Shaking [$0^{\circ},15^{\circ}$], Rolling [$0^{\circ},45^{\circ}$], and Tipping [$0^{\circ},90^{\circ}$], where Horizontal corresponds to the standard MDE evaluation setting. Since roll angles beyond $90^{\circ}$ are extremely rare and can be losslessly mapped back to this valid range via a $90^{\circ}$ rotation, our experiments focus on these four representative intervals. Experimental results demonstrate that both ViT-based and diffusion-based MDE models suffer from substantial performance degradation as the horizontal roll increases, highlighting the severity of the horizontal prior. Although two intuitive solutions can only partially mitigate this issue, our proposed ID-Constraint achieves consistently better results on non-horizontal tail cases, significantly enhancing model robustness across diverse orientations. 


The main contributions of this paper are threefold: 1) we present the first systematic investigation of the Horizontal Prior and demonstrate that it arises from a long-tailed distributional bias in MDE; 2) we propose ID-Constraint to mitigate the Horizontal Prior through roll-robust supervision during training; 3) we establish a new benchmark for distribution-aware depth estimation, comprising 5 datasets evaluated under 4 distinct roll settings; 4) extensive experiments demonstrate that our approach outperforms previous state-of-the-art MDE methods in terms of rolling robustness.

\section{Related Work}
\label{sec:related}

\noindent\textbf{Monocular depth estimation.}
MDE~\cite{zhao2020monocular,masoumian2022monocular,dav3} is a fundamental task that aims to predict the depth value of each pixel from a single image. Early approaches can be broadly categorized into three groups. The first group relies on hand-crafted features and geometric priors for depth inference~\cite{sturm1996factorization,zhang2025review} but often suffer from poor generalization in complex, real-world environments. The second group employs CNN-based end-to-end training, which has achieves significant progress but remains constrained by the availability of large-scale, high-quality annotated data~\cite{yao2020blendedmvs,cho2021diml}. The third group enhances model performance by incorporating auxiliary supervision such as semantic segmentation and surface normal prediction~\cite{ke2023marigold,eigen2015predicting}.
Recent advances in visual foundation models, including self-supervised vision backbones~\cite{oquab2024dinov2,he2020momentum} and diffusion-based architectures~\cite{Rombach_2022_CVPR,peebles2023scalable}, have driven a new wave of advances in MDE~\cite{yang2024depth,birkl2023midas,ke2023marigold}. In particular, Depth Anything V2 (DAv2)~\cite{yang2024depthv2}, supported by large-scale high-quality synthetic datasets~\cite{yao2020blendedmvs,wang2020tartanair}, has pushed depth fidelity to unprecedented levels. However, despite these remarkable achievements, existing MDE models remain vulnerable to small image perturbations, revealing a persistent robustness challenge in MDE.

\noindent\textbf{Long-tailed distribution.}
Research on long-tailed distributions has primarily focused on classification tasks~\cite{zhang2023deep,kang2019decoupling,tang2020long}. In such settings, object occurrence naturally follows a long-tailed distribution~\cite{liu2019large}, resulting in significant disparities in sample density and learning difficulty across classes. Consequently, models trained on these imbalanced datasets typically achieve lower accuracy and recall on tail (rare) classes.
Conventional approaches to long-tailed classification can be broadly categorized into two groups: data augmentation~\cite{hu2020learning,tang2022invariant} and distribution adjustment~\cite{tang2020long,menon2020long}. Data augmentation methods aim to re-balance the class distribution during training, often by re-sampling~\cite{kim2020m2m} or generating synthetic tail-class samples~\cite{yin2019feature}. In contrast, distribution adjustment methods adopt training-free strategies that calibrate model predictions during inference~\cite{menon2020long,zhang2022self}, thereby improving recognition of under-represented classes. In MDE, prior studies on data distribution have primarily examined the depth distribution~\cite{yu2024dme,zhan2025vistadepth}. To the best of our knowledge, our work is the first to identify the horizontal prior issue, revealing a new source of long-tailed bias in MDE.

\section{Method}
\label{sec:method}


\subsection{Preliminaries and Baseline Methods}
An MDE model aims to predict a dense depth map $D^p \in \mathbb{R}^{H \times W}$ from a single image $I \in \mathbb{R}^{C \times H \times W}$, where $C$ denotes the number of RGB channels, $H$ and $W$ represent the height and width of the input image, respectively. Recent works can be broadly categorized into diffusion-based generative models~\cite{ke2023marigold,xu2025genpercept} and ViT-based feedforward models~\cite{yang2024depthv2,he2025distill}. The latter, represented by Depth Anything V2 (DAv2)~\cite{yang2024depthv2}, currently demonstrates superior performance. Therefore, our algorithms are developed based on ViT-based MDE models that employ a fine-tuned DINOv2~\cite{oquab2024dinov2,dosovitskiy2020image} backbone to extract multi-level feature maps $\{F_i\} = \text{ViT}(I)$, where each $F_i \in \mathbb{R}^{C_i \times \frac{H}{p} \times \frac{W}{p}}$ denotes the output feature map at level $i$ from the ViT encoder, $C_i$ is the number of channels, and $p$ represents the patch size. A DPT decoder head~\cite{ranftl2021vision} is then applied to these feature maps to directly generate the final depth map, $D^p = \text{DPT}(\{F_i\})$.

To accelerate the training process, we adopt the distillation pipeline from Distill Any Depth (DistillAD)~\cite{he2025distill} as our baseline. This approach requires fewer data and less training iterations while achieving performance comparable to DAv2~\cite{yang2024depthv2}. The distillation loss is defined as follows:
\begin{equation}
    \mathcal{L}_{distill} = \frac{1}{M} \sum_{i=1}^M \left| \mathcal{N}(D^p)_i -  \mathcal{N}(D^{t})_i \right|, \\
\end{equation}
where $\mathcal{N}(\cdot)$ denotes the normalization function, $M$ is the number of valid pixels, and $D^{t}$ represents the depth map predicted by the teacher model, \textit{i.e.}, DAv2 in our experiments. In this paper, we adopt global normalization as the normalization strategy used in $\mathcal{L}_{distill}$, which is defined as $\mathcal{N}(D) = \frac{D - med(D)}{\frac{1}{M} \sum_{i=1}^M \left| D_i - med(D) \right|}$, where $med(\cdot)$ denotes the median value of all valid pixels in a depth map. 

We also adopt the gradient matching loss $\mathcal{L}_{gm}$ used in MiDaS~\cite{Ranftl2022,birkl2023midas} and DAv2, making the final loss function for our re-implemented baseline $\mathcal{L} = \mathcal{L}_{distill} + \mathcal{L}_{gm}$. Following the widely adopted scale- and shift-invariant formulation~\cite{yang2024depthv2}, the predicted depth map $D^{p}$ is transformed into an affine-invariant inverse depth space prior to loss computation. Note that DAv2 and DistillAD have not released their training and evaluation code, so we have to re-implement both methods.

\begin{figure}[tb]
    \centering
    \includegraphics[width=1.0\linewidth]{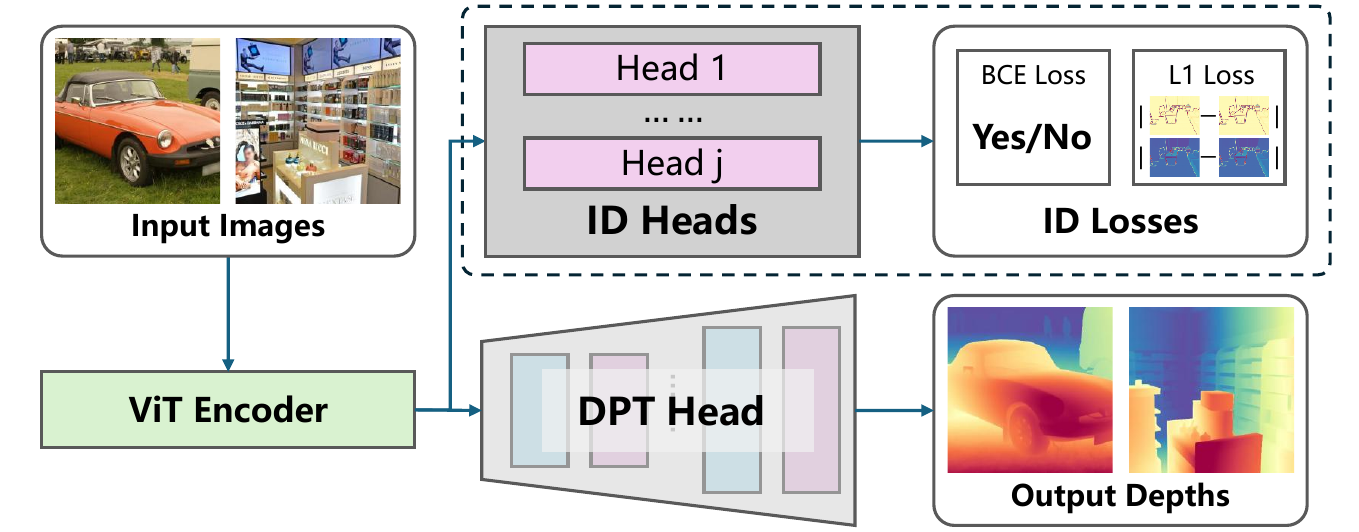}
    \caption{The proposed ID-Constraint with auxiliary ID Heads to enhance the roll-robust feature learning.}
    \label{fig3:method}
\end{figure}

\subsection{The Horizontal Prior}

\noindent\textbf{Definition.} The horizontal prior refers to the phenomenon whereby naturally collected real-world images tend to be captured in approximately horizontal orientations, resulting in a long-tailed distribution of absolute roll angles, as illustrated in Figure~\ref{fig1:longtail}. Consequently, the depth estimation error for a rolled image $I_{\theta}$ with the roll angle $\theta > 0^{\circ}$ is significantly larger than that of the corresponding horizontal image $I_{0}$, as evidenced by the experimental results in Table~\ref{tab1:main}.

To address this robustness challenge, two intuitive approaches can be considered: (1) re-balanced augmentation and (2) horizon leveling:

\textbf{1) Re-balanced augmentation.} Following the re-balanced augmentation in long-tailed classification tasks~\cite{hu2020learning}, we can directly augment each image by uniformly sampling an angle $\theta \in [-90^\circ, 90^\circ]$ and rotating the image accordingly in MDE. To prevent the model from learning shortcut cues from padded boundaries, a random center crop is simultaneously applied, uniformly preserving from 40\% to 100\% of the original image size.

\textbf{2) Horizon leveling.} Commercial horizon-leveling solutions typically rely on hardware sensors to estimate the camera rolling angle, and then rotate and crop the image to restore horizontal alignment. However, such sensor measurements are unavailable for large-scale image datasets. We therefore estimate the roll angle directly from the image. We develop a two-stage training framework with data denoising. In stage one, we treat most images as horizontally aligned due to horizontal prior and use the applied rotation angle $\theta$ as supervision. Since those originally tilted images introduce label noise, we have errors above $45.69^\circ$ at this stage on our test set. Inspired by prior denoising methods~\cite{northcutt2021confident,yi2022identifying}, we assume that samples with larger prediction errors are more likely to be non-horizontal. We therefore remove half of highest-error samples and retrain on the cleaned subset as stage two. As shown in Table~\ref{tab4:angle}, denoised data substantially improves most models, reducing the best error to $25.90^\circ$. Finally, each image is rotated by its predicted roll angle to restore horizontal alignment.

\begin{table}[t]
\centering
\fontsize{7}{9}\selectfont
\begin{tabular}{cccccccccccccc} 
\hline
\toprule
 Stage Number & Activation & L1 loss & COS loss & Angle Error~($^{\circ}$) \\

\midrule

Stage-1 & tanh & \ding{51} & & 45.69 \\
Stage-1 & tanh & & \ding{51} & 47.94 \\

\midrule

Stage-2 & none & \ding{51} & \ding{51} & 28.53\\
Stage-2 & none & & \ding{51} & 40.41\\
Stage-2 & none & \ding{51} & & 46.32\\
Stage-2 & tanh & \ding{51} & \ding{51} & 29.49\\
Stage-2 & tanh & & \ding{51} & 31.99\\
Stage-2 & tanh & \ding{51} &  & \textbf{25.90}\\

\bottomrule
\hline
\end{tabular}
\caption{Experimental results of roll-angle prediction models under different configurations.}
\label{tab4:angle}
\end{table}
\noindent\textbf{Limitations of intuitive approaches.} 
However, both intuitive remedies have clear limitations. Re-balanced augmentation can disrupt feature learning. Because DINOv2 was pretrained without such augmentation, it may enlarge the domain gap between pretraining and finetuning data. Meanwhile, horizon leveling is limited by inaccurate roll-angle estimation: even mild rotations of $[0^\circ,15^\circ]$ cause substantial degradation, while the best angle predictor still has an average error of $25.90^\circ$.


\subsection{Invariant Depth Constraint}
\label{sec:3.3-idcue}
To address the above limitations, we propose the Invariant Depth Constraint (ID-Constraint), which encourage the model to learn depth-relevant features that remain invariant under rotation. Inspired by \citet{danier2025depthcues}, which shows that advances in MDE largely arise from implicitly learning human-like depth cues such as elevation, occlusion, and perspective, we note that several of these tasks are inherently rotation-invariant and can therefore mitigate the horizontal prior. Accordingly, we adopt four region-level supervision tasks and further introduce two pixel-level supervision tasks as invariant depth constraints for fine-grained feature learning. Detailed examples are provided in the Appendix.


\noindent\textbf{Region-level constraints.} We select four region-level tasks from the original DepthCues benchmark~\cite{danier2025depthcues} whose predictions remain consistent under rotation, \textit{i.e.}, they are rotation-invariant: \textbf{(1) Light and shadow}, which requires identifying whether a shadow belongs to an object; \textbf{(2) Occlusion}, which determines whether an object is partially blocked or not; \textbf{(3) Size}, which infers the relative size of two objects; and \textbf{(4) Texture gradient}, which estimates the relative depth of two regions from the degree of texture compression. All these tasks are formulated as binary classifications constraints. Each task is trained with a BCE loss and predicted by $Y_j = \text{Head}_j(\text{ViT}(I_\theta), \{M_k\})$, where $Y_j$ denotes the output of task $j$, $\text{Head}_j(\cdot)$ represents the corresponding prediction head following \cite{danier2025depthcues}, $\{M_k\}$ represents the object masks, which can be either a single-object mask $M_a$ or a dual-object pair $\{M_a, M_b\}$, and $\theta$ is a random augmentation angle for input image $I_\theta$.

\noindent\textbf{Pixel-level constraints.} As shown in Figure~\ref{fig4:visual}, MDE models exhibit substantially degraded local geometric understanding on rolled images. We therefore introduce two pixel-level tasks, \textbf{(5) Local Peak} and \textbf{(6) Local Slope}, to capture rotation-invariant local geometry and support fine-grained feature learning. The local peak map measures convex and concave structures by subtracting each pixel’s depth from the mean valid depth within a $5\times5$ neighborhood, implemented with a handcrafted convolution kernel. The local slope map computes the mean absolute depth gradient along four directions. These two maps are formulated as follows, with visual examples provided in the Appendix:
\begin{align}
    Z^{p} &= (D^t \otimes K_1) / (M^{vp} \otimes K_1) - D^t, \label{eq:peak}\\
    Z^{s} &= \frac{1}{4} \sum_d(\left| D^t \otimes K_d \right|), \label{eq:slope}
\end{align}
where $Z^{p}, Z^{s} \in \mathbb{R}^{H \times W}$ denote the local peak and local slope maps, respectively, $\otimes$ represents the convolution operation, $K_1$ is a $5 \times 5$ kernel filled with ones, $K_d \in \{K_\downarrow, K_\rightarrow, K_\searrow, K_\swarrow\}$ are $5 \times 5$ directional slope-detection convolution kernels, and $M^{vp} \in \{0,1\}^{H \times W}$ denotes the valid-pixel mask. A shared DPT head, $Y_{pixel} = \text{Head}_{pixel}(\text{ViT}(I_\theta))$, serves as the prediction head for both tasks, and then supervised by $\{Z^p_\theta, Z^s_\theta\}$ using L1 loss, where $Z^p_\theta$ and $Z^s_\theta$ denote the rotated versions of the horizontal peak and slope maps $Z^p$ and $Z^s$ computed from $I_0$.




\noindent\textbf{Training Strategy.} 
Building on the auxiliary tasks described above, we fine-tune the DAv2 backbone with a set of Invariant Depth heads and corresponding losses, termed ID Heads and ID Losses, as illustrated in Figure~\ref{fig3:method}. During training, the ViT encoder is jointly optimized using the ID and MDE losses to promote roll-invariant feature representations. At inference, the ID Heads are discarded, preserving the enhanced robustness without introducing additional computational overhead.



\section{Experiments}
\label{sec:exp}

\begin{table*}[t]
\centering
\fontsize{10}{10}\selectfont
\begin{tabular}{lccccccccccccc} 
\hline
\toprule
\multirow{2}{*}{\textbf{Method}} 
          &  & \multicolumn{2}{c}{\textbf{Horizontal ($0^{\circ}$)}} 
          &  & \multicolumn{2}{c}{\textbf{Shaking [$0^{\circ},15^{\circ}$]}} 
          &  & \multicolumn{2}{c}{\textbf{Rolling [$0^{\circ},45^{\circ}$]}} 
          &  & \multicolumn{2}{c}{\textbf{Tipping [$0^{\circ},90^{\circ}$]}} 
          \\ 
    \cmidrule{3-4}\cmidrule{6-7}\cmidrule{9-10}\cmidrule{12-13}

 & & AbsRel$\downarrow$ & $\delta_1\uparrow$ & & AbsRel$\downarrow$ & $\delta_1\uparrow$ &
   & AbsRel$\downarrow$ & $\delta_1\uparrow$ & & AbsRel$\downarrow$ & $\delta_1\uparrow$ \\

\midrule

Marigold\textsuperscript{*}~\cite{ke2023marigold} & & 0.130 & 0.886 & & 0.146 & 0.859 & & 0.171 & 0.814 & & 0.200 & 0.765 \\
GenPercept\textsuperscript{*}~\cite{xu2025genpercept} & & 0.130 & 0.891 & & 0.143 & 0.869 & & 0.164 & 0.830 & & 0.185 & 0.794 \\

\midrule

DAv2\textsuperscript{*}~\cite{yang2024depthv2} & & \textbf{0.096} & \textbf{0.922} & & 0.109 & 0.906 & & 0.119 & 0.895 & & 0.127 & 0.883 \\
DistillAD\textsuperscript{*}~\cite{he2025distill} & & \underline{0.099} & \underline{0.920} & & 0.113 & 0.900 & & 0.124 & 0.889 & &	0.131 & 0.878 \\

\midrule

(ours) Baseline & & 0.100 & \underline{0.920} & & 0.113 & 0.901 & & 0.125 & 0.887 & & 0.135 & 0.873 \\
(ours) Re-balanced Aug & & 0.104 & 0.918 & & 0.111 & 0.904 & & 0.114 & 0.901 & & 0.119 & 0.897 \\

(ours) Horizon Leveling & & 0.100 & \underline{0.920} & & \underline{0.108} & \underline{0.910} & & \underline{0.110} & \underline{0.907} & & \underline{0.112} & \underline{0.905} \\ 
(ours) ID-Constraint &  & \underline{0.099} & \underline{0.920} & & \textbf{0.103} & \textbf{0.916} &  & \textbf{0.104} & \textbf{0.915} &  & \textbf{0.106} & \textbf{0.913} \\

\bottomrule
\hline
\end{tabular}
\caption{Experimental results of state-of-the-art MDE models and the proposed algorithms under four roll settings. Results for each setting are averaged across five benchmarks. Superscript \textsuperscript{*} indicates results evaluated using our reimplemented evaluation code. $\downarrow$ denotes lower values are better, while $\uparrow$ denotes higher values are better. \textbf{Bold} and \underline{underline} indicates the best and second best results, respectively.}
\label{tab1:main}
\end{table*}


\subsection{Implementation Details}

\noindent\textbf{Training Details.} Following DistillAD~\cite{he2025distill}, we train our MDE baselines on 200,000 unlabeled real-world images from the SA-1B~\cite{kirillov2023segany} dataset. For ID-Constraint, we sample 20,000 images from SA-1B as pixel-level constraint data. As to the region-level constraint tasks, we collect 4,716 training samples for light and shadow, 24,402 samples for occlusion, and 1,986 and 4,000 samples for size and texture gradient, respectively, following the original DepthCues benchmark~\cite{danier2025depthcues}. The MDE losses are applied to all samples described above. We use the state-of-the-art MDE model DAv2~\cite{yang2024depthv2} as the teacher to produce pseudo-depth labels for horizontal images. We also adopt the same input pre-processing strategy from DistillAD, where each input image is resized and cropped to $560 \times 560$ and then normalized before prediction. We utilize the AdamW~\cite{loshchilov2017decoupled} optimizer with different learning rates for ViT encoder ($5 \times 10^{-6}$) and ID heads or DPT head ($5 \times 10^{-5}$). The DPT head is re-initialized before MDE distillation training. We train all MDE models and ID-Constraint models for 1 epoch with a batch size of 8, and roll-angle prediction models for 5 epochs with a batch size of 64. The ID losses are directly added to the final loss with weight 1.0. For roll-angle prediction models, the training loss is L1 loss or cosine similarity (COS) loss. 

\noindent\textbf{Model details.} For MDE models, we adopt DAv2 with ViT-Large encoder as the distillation teacher, and the student model chooses the same encoder and outputs 4 feature maps at layers 4, 11, 17, and 23. For ID-Constraint finetuning, all region-level prediction heads are implemented following \cite{danier2025depthcues} and pixel-level prediction head is a DPT head with smaller hidden dimension 128. The roll-angle prediction models use a ViT-Small backbone with a prediction head composed of an attention pooling, a 1D batch norm, two linear layers, and an optional $tanh$ activation. Instead of directly predicting angles, we predict a 2D vector, $cos(\theta)$ and $sin(\theta)$, as they are inherently normalized. The rolling angle $\theta$ can be converted back from the predicted 2D vector. 


\subsection{Evaluation Settings}
We evaluate all models on five benchmark datasets: DIODE~\cite{vasiljevic2019diode}, ScanNet~\cite{dai2017scannet}, ETH3D~\cite{schops2017multi}, KITTI~\cite{geiger2012we}, and NYUv2~\cite{silberman2012indoor}. During implementation, we observed that each test dataset requires a distinct pre-processing step to determine valid pixels. However, neither DAv2~\cite{yang2024depthv2} nor DistillAD~\cite{he2025distill} have released their evaluation code. Therefore, our evaluation pipeline is based on GenPercept~\cite{xu2025genpercept}, which may introduce minor inconsistencies between our reproduced results and those reported in the original papers. Furthermore, both DAv2 and DistillAD, as well as our models, are evaluated and produce outputs in the disparity (inverse-depth) space, whereas Marigold~\cite{ke2023marigold} and GenPercept~\cite{xu2025genpercept} operate in the depth space. Following prior work~\cite{yang2024depthv2,he2025distill,xu2025genpercept}, the predicted depth maps adopt scale- and shift-invariance transformation before evaluation. 

To better evaluate the robustness degradation induced by the horizontal prior, we define four roll settings: Horizontal ($0^{\circ}$), Shaking [$0^{\circ},15^{\circ}$], Rolling [$0^{\circ},45^{\circ}$], and Tipping [$0^{\circ},90^{\circ}$]. Among them, Horizontal ($0^{\circ}$) corresponds to the conventional test setting without manual rotation, while Shaking, Rolling, and Tipping represent three rolling levels, where each image is rotated by an angle uniformly sampled within the specified range. For fair comparisons, the random seed is fixed across all evaluations. For roll-angle prediction evaluation, we use the same five benchmark datasets under the Tipping scenario. For all depth evaluations, we adopt two commonly used metrics for quantitative assessment: mean absolute relative error (AbsRel) and $\delta_1$ accuracy, where lower AbsRel and higher $\delta_1$ indicate better performance.

\begin{table*}[t]
\centering
\fontsize{8}{10}\selectfont
\begin{tabular}{lccccccccccccccc} 
\hline
\toprule
\multirow{2}{*}{\textbf{Method}} 
          & \multicolumn{2}{c}{DIODE}
          &  & \multicolumn{2}{c}{ScanNet}
          &  & \multicolumn{2}{c}{ETH3d}
          &  & \multicolumn{2}{c}{KITTI}
          &  & \multicolumn{2}{c}{NYUv2}
          \\ 
    \cmidrule{2-3}\cmidrule{5-6}\cmidrule{8-9}\cmidrule{11-12} \cmidrule{14-15}

   & AbsRel$\downarrow$ & $\delta_1\uparrow$ & & AbsRel$\downarrow$ & $\delta_1\uparrow$ &
   & AbsRel$\downarrow$ & $\delta_1\uparrow$ & & AbsRel$\downarrow$ & $\delta_1\uparrow$ & 
   & AbsRel$\downarrow$ & $\delta_1\uparrow$ \\

\midrule

Marigold\textsuperscript{*}~\cite{ke2023marigold} & 0.337 & 0.705 & & 0.133 & 0.832 & & 0.129 & 0.842 & & 0.246 & 0.613 & & 0.125 & 0.854 \\
GenPercept\textsuperscript{*}~\cite{xu2025genpercept} & 0.337 & 0.715 & & 0.113 & 0.871 & & 0.114 & 0.877 & & 0.225 & 0.633 & & 0.103 & 0.895 \\

\midrule

DAv2\textsuperscript{*}~\cite{yang2024depthv2} & 0.268 & 0.736 & & 0.078 & 0.938 & & 0.075 & 0.941 & & 0.118 & 0.873 & & 0.069 & 0.957 \\
DistillAD\textsuperscript{*}~\cite{he2025distill} & 0.271 & 0.735 & & 0.079 & 0.939 & & 0.080 & 0.932 & & 0.124 & 0.863 & & 0.074 & 0.952 \\

\midrule

(ours) Baseline & 0.279 & 0.732 & & 0.085 & 0.923 & & 0.082 & 0.929 & & 0.124 & 0.863 & & 0.074 & 0.950 \\
(ours) Re-balanced Aug & 0.263 & \underline{0.748} & & \underline{0.058} & \underline{0.967} & & 0.068 & 0.957 & & 0.115 & 0.877 & & 0.062 & \underline{0.967} \\
(ours) Horizon Leveling & \underline{0.262} & 0.745 & & 0.061 & 0.958 & & \underline{0.057} & \underline{0.967} & & \underline{0.087} & \underline{0.927} & & \underline{0.061} & 0.962 \\
(ours) ID-Constraint & \textbf{0.255} & \textbf{0.753} &  & \textbf{0.051} & \textbf{0.971} &  & \textbf{0.051} & \textbf{0.972} &  & \textbf{0.086} & \textbf{0.931} &  & \textbf{0.055} & \textbf{0.971} \\

\bottomrule
\hline
\end{tabular}
\caption{Experimental results of state-of-the-art MDE models and the proposed algorithms under the hardest Tipping [$0^{\circ},90^{\circ}$] setting. Superscript \textsuperscript{*} indicates results evaluated using our reimplemented evaluation code.}
\label{tab2:5datasets}
\end{table*}

\subsection{Experimental results}

\begin{figure*}[tb]
    \centering
    \includegraphics[width=1.0\linewidth]{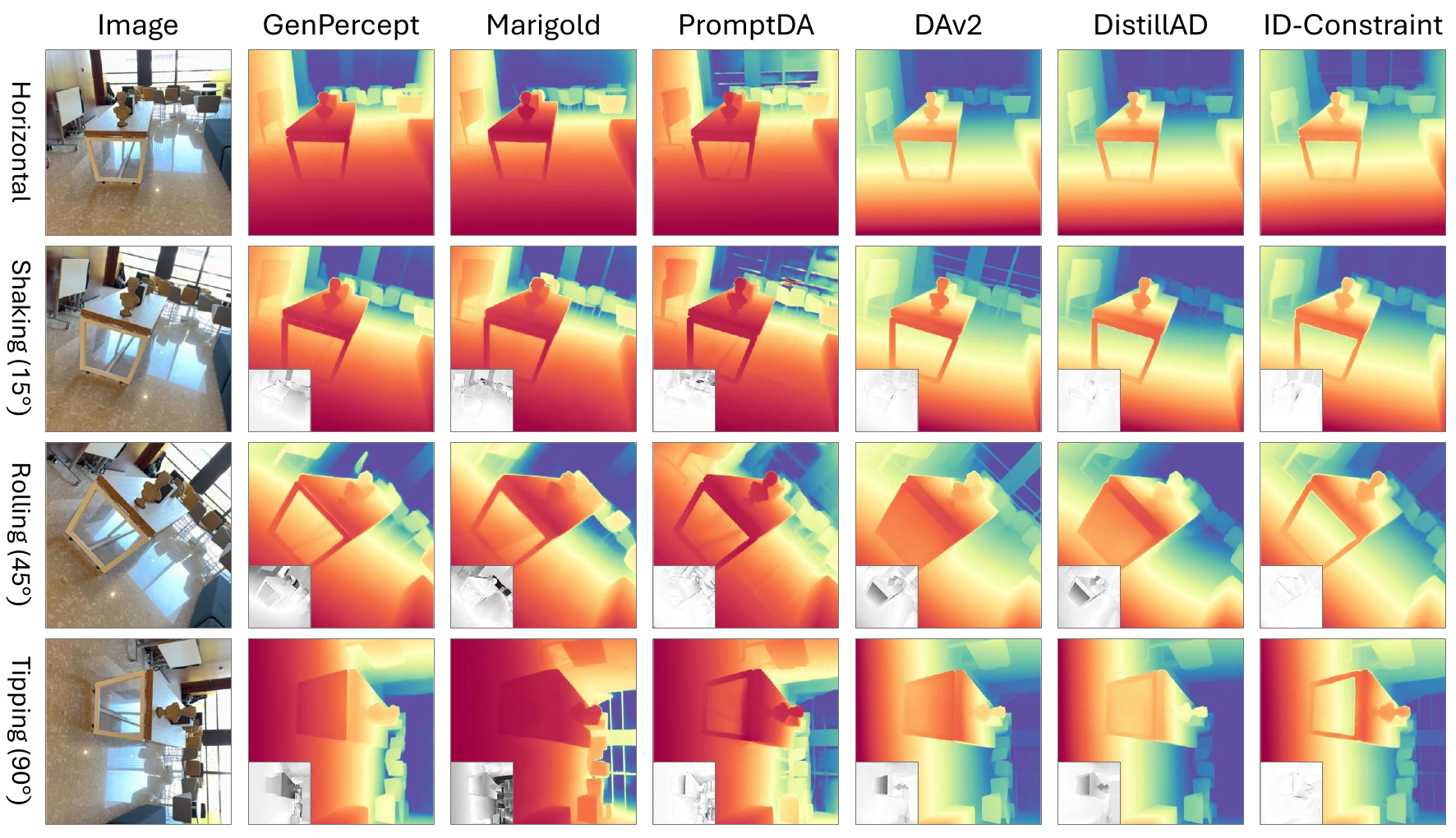}
    \caption{Qualitative comparisons of existing state-of-the-art MDE models and the proposed and ID-Constraint, on a real-world indoor image under four roll settings. Note that PromptDA takes an additional low-resolution ground-truth depth map as input. The bottom-left grayscale images visualize the absolute error maps corresponding to their respective horizontal depth predictions. GenPercept, Marigold, and PromptDA produce outputs in the depth space, while the remaining output in the disparity space.}
    \label{fig4:visual}
\end{figure*}

\noindent\textbf{Comparisons with state-of-the-art methods.} Existing state-of-the-art (SOTA) MDE methods can be broadly categorized into two groups: diffusion-based and ViT-based approaches. In our experiments, we evaluate four representative models: Marigold~\cite{ke2023marigold} and GenPercept~\cite{xu2025genpercept} as diffusion-based methods, and DAv2~\cite{yang2024depthv2} and DistillAD~\cite{he2025distill} as ViT-based feedforward methods. 
As shown in Table~\ref{tab1:main}, all four existing SOTA methods suffer from the long-tailed bias introduced by the horizontal prior, with their performance consistently degrading as the roll range increases. In contrast, both intuitive solutions and ID-Constraint approaches exhibit robustness against this bias. Among them, ID-Constraint achieves the best performance across the Shaking, Rolling, and Tipping settings. Its slightly lower accuracy than DAv2 in the Horizontal setting is primarily attributed to the weaker performance of our reimplemented baseline. 
For individual benchmark, we report detailed results on all five datasets under the Tipping scenario in Table~\ref{tab2:5datasets}, and the observations remain consistent. The proposed ID-Constraint outperforms existing SOTA methods and other solutions in both metrics.
To better illustrate the robustness issues introduced by the horizontal prior, we present an example of MDE predictions from all methods under the four roll settings in Figure~\ref{fig4:visual}. All existing SOTA models fail to accurately recover the depth around the table legs as the rolling angle increases. It is worth noting that PromptDA~\cite{lin2025promptda} is not an MDE method, because it requires a low-resolution ground-truth depth map as input, making it a depth super-resolution model. Nevertheless, even with this model, the table boundaries still become noticeably blurred in the Tipping condition, further highlighting the challenge posed by the horizontal prior.

\begin{table}[t]
  \centering
  \fontsize{6}{9}\selectfont
  \begin{tabular}{ccccccc} 
  \hline
  \toprule
   \makecell{{Re-balanced}\\{Aug}}  
& \makecell{{Horizon}\\{Leveling}}
& \makecell{Region\\Losses}
& \makecell{Pixel\\Losses} 
& \makecell{{ID-}\\{Constraint}} 
& AbsRel$\downarrow$ & $\delta_1\uparrow$ \\

\midrule

& & & & & 0.135 & 0.873 \\

\midrule

\ding{51} & & & & & 0.119 & 0.897 \\
\ding{51} &  & \ding{51} & \ding{51} & \ding{51} & 0.114 & 0.903 \\

\midrule

& \ding{51} & & & & 0.112 & 0.905 \\
\ding{51} & \ding{51} & & & & 0.110 & 0.909 \\

\midrule

\ding{51} & \ding{51} &  \ding{51} & & \ding{51} & 0.108 & 0.910 \\
\ding{51} & \ding{51} &  & \ding{51} & \ding{51} & 0.107 & 0.912 \\

\ding{51} & \ding{51} & \ding{51} & \ding{51} & \ding{51} & \textbf{0.106} & \textbf{0.913} \\

   \bottomrule
   \hline
   \end{tabular}
   \caption{Ablation study analyzing the impact of each module under the most challenging Tipping setting. Reported results are averaged across five benchmarks.}
\label{tab:ablation}
\end{table}


\noindent\textbf{Ablation studies.} To assess the impact of the re-balanced augmentation, horizon leveling, and the region-/pixel-level losses in the proposed ID-Constraint, we conduct a comprehensive ablation study under the most challenging Tipping [$0^{\circ}, 90^{\circ}$] setting in Table~\ref{tab:ablation}. 
Since ID-Constraint requires re-balanced augmentation for invariant feature learning, re-balanced augmentation is used as the default setting for ID-Constraint. 
Meanwhile, horizon leveling can be regarded as a training-free adjustment, and can be applied on top of all previous settings to further enhance the robustness of the models. As to the region and pixel constraints, we conduct ablation study on each group, both components improve performance over the baseline. In particular, pixel constraints provide a higher performance gain when individually evaluated. Note that we did not exhaustively evaluate all possible 6 task combinations, as this would result in $2^6-1=63$ distinct experimental settings.



\begin{figure}[t]
\centering

\includegraphics[width=\linewidth]{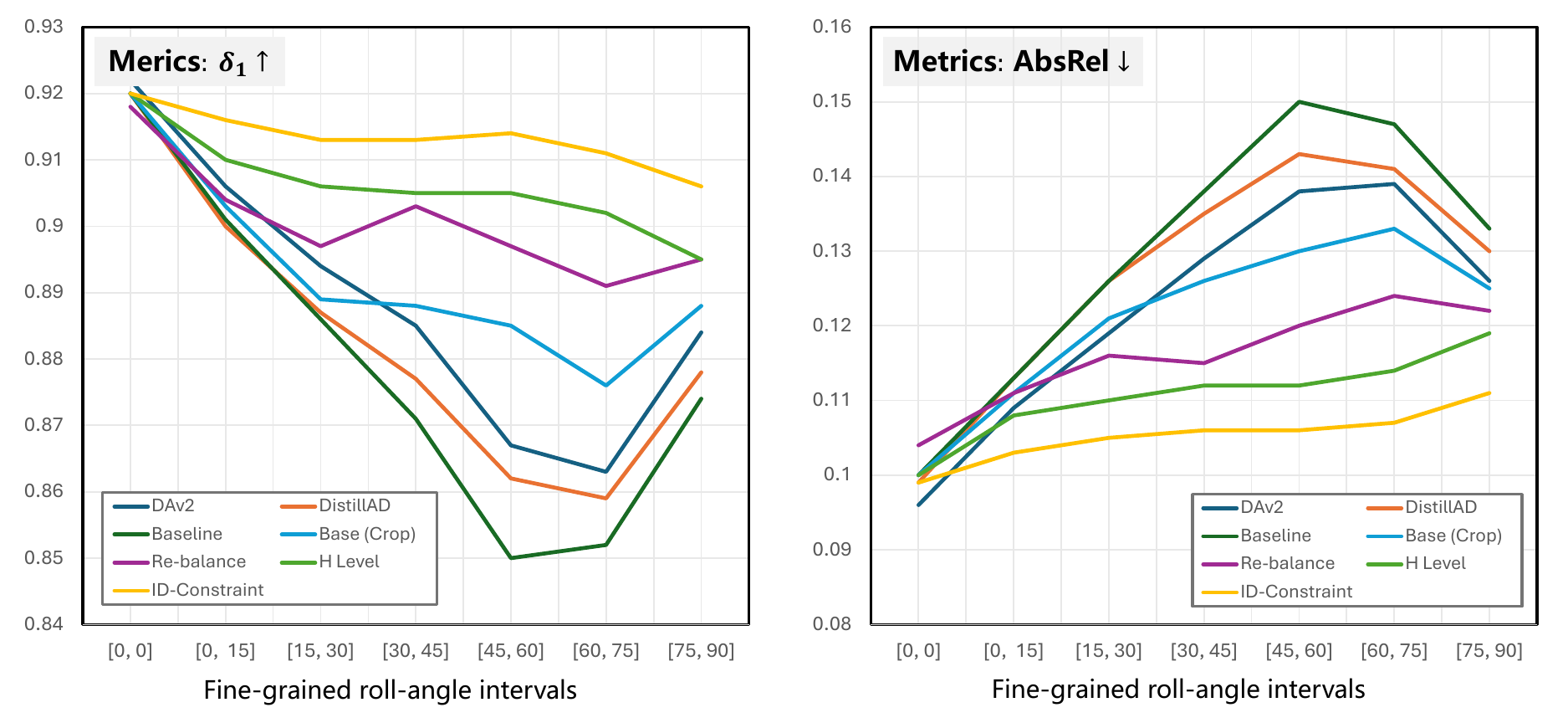}
  \captionof{figure}{The averaged results across five benchmarks on fine-grained roll intervals. H Level means horizon leveling.}
  \label{fig5:interval}

\end{figure}

\subsection{Further analyses}

To better understand the horizontal prior problem, we also conduct the following additional analyses.

\noindent\textbf{The prevalence of the horizontal prior problem.} Since the horizontal prior problem originates at the data level, both diffusion-based pipelines and ViT-based feedforward pipelines are affected. As a result, as shown in Figure~\ref{fig4:visual}, all previous MDE models fail to distinguish the table legs under severe Tipping conditions. Interestingly, even PromptDA~\cite{lin2025promptda}, which takes a low-resolution ground-truth depth map as input, still produces noticeably blurred edges around the table legs, suggesting that the horizontal prior problem can even affect non-MDE models.

\noindent\textbf{Investigating more fine-grained roll-angle intervals.} To analyze model performance under finer roll-angle intervals, we plot the averaged results at every $15^{\circ}$ interval across the five benchmark datasets in Figure~\ref{fig5:interval}. Within the range of [$0^{\circ},60^{\circ}$], model performance gradually degrades as the roll angle increases, but an unexpected improvement appears near $90^{\circ}$. This rebound occurs because a small portion of photos taken by cameras or smartphones fail to undergo horizontal correction. This phenomenon does not appear in Figure~\ref{fig1:longtail}(d), because we manually rotate those images with roll angles greater than $45^{\circ}$ for alignment.

\noindent\textbf{The impact of image resolution.} In general, since the ViT input is resized so that its shorter side becomes 560, larger original images tend to produce less accurate depths. Therefore, as the image rotates from the horizontal position to $45^{\circ}$, its effective resolution gradually increases, which also leads to performance degradation. To verify that the horizontal prior problem is not solely an artifact of resolution changes, we introduce a Base (Crop) setting in Figure~\ref{fig5:interval}, where each rotated image is cropped to match the original resolution. As shown, although the performance improves slightly with cropping, the overall downward trend remains consistent with that of the baseline.

\noindent\textbf{Real-world deployment and practical significance.} Our investigation into the Horizontal Prior problem was motivated by a practical downstream applications (Figure~\ref{fig2:case}). Specifically, in a mobile-captured real-world indoor scene reconstruction project, we often observed blurred boundaries. Our further diagnostic analysis reveals that these artifacts arise from inaccurate depth, because existing MDE models (\textit{e.g.}, DAv2 in our previous framework) have inherent sensitivity to roll-angle perturbations, which can be alleviated by the proposed method.

\section{Conclusion}
\label{sec:conclusion}

In this paper, we identify the horizontal prior problem in MDE for the first time, providing a new perspective on how long-tailed data distributions introduce bias in depth estimation. We explore two intuitive solutions, re-balanced augmentation and horizon leveling, which partially alleviate this issue. To further enhance robustness, we propose the Invariant Depth Constraint (ID-Constraint), a feature-learning framework that applies four region-level and two pixel-level supervision losses with prediction heads to the ViT encoder. These auxiliary heads are discarded at inference, introducing no additional computational cost. ID-Constraint achieves the best overall performance across five benchmarks and three out of four roll settings. We believe that our work sheds new light on the robustness of MDE and offers new insights into generalization research across other related tasks.

\section{Acknowledgement}
\label{sec:acknowledgement}

This work was supported by the Fundamental Research Funds for the Central Universities at Tongji University under Grant No.~22120260376.

\bibliography{aaai2027}

\appendix


\appendix

\section{Appendix}
This supplementary material provides the following additional information: (B) extended dataset details; (C) a more in-depth analysis of the long-tailed roll angle distribution; (D) additional experimental details and results; and (E) additional qualitative visualizations.

\section{Extended Dataset Details}
\subsection{Training Datasets}
In this paper, all MDE models are trained on 200,000 unlabeled SA-1B~\cite{kirillov2023segany} images using pseudo depth labels generated by DAv2~\cite{yang2024depthv2}. Invariant Depth Constraint (ID-Constraint) are trained with additional a subset of the DepthCues benchmark~\cite{danier2025depthcues} together with 20,000 SA-1B samples. For those 200,000 MDE training data, Invariant Depth Losses (ID Losses) are removed with weight $0.0$. Detailed configurations and procedures of each training dataset are provided below:

\textbf{1) SA-1B~\cite{kirillov2023segany}}: The full SA-1B dataset contains 11M diverse real-world images, distributed across 1,000 zipped files. Due to computational constraints, we follow DistillAD~\cite{he2025distill} to sample 20 files, sa\_000000.tar, sa\_000050.tar, sa\_000100.tar, ..., sa\_000950.tar, resulting in 200,000 training images. For baseline training, we apply only random cropping and RGB normalization. For re-balanced angle augmentation, we uniformly sample an augmentation angle from $-90^{\circ}$ to $90^{\circ}$, while keeping 10\% none augmented samples during training.
    
\begin{figure}[tb]
    \centering
    \includegraphics[width=1.0\linewidth]{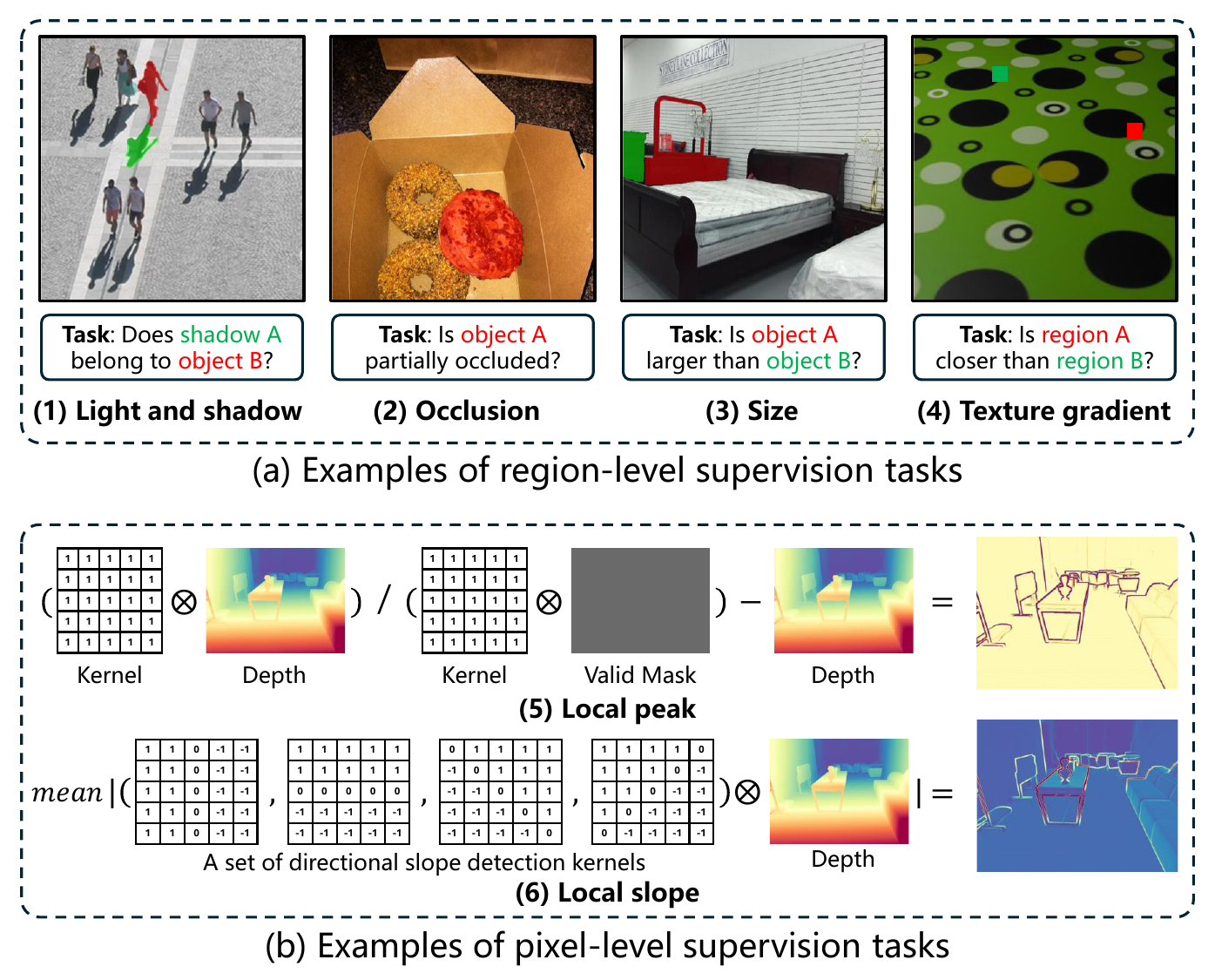}
    \caption{(a) Examples of four region-level constraint supervision tasks whose prediction labels remain consistent under rotation. (b) Examples of two pixel-level constraint supervision tasks that compute rotation-invariant local peaks of depth and average absolute slopes in 4 directions.}
    \label{fig2:idcue}
\end{figure}

\textbf{1) DepthCues~\cite{danier2025depthcues}}: The original DepthCues benchmark includes six task settings. For training our proposed ID-Constraint, we use four of them: light–shadow, occlusion, size, and texture–gradient. Because these subsets have different data scales, we upsample the light–shadow, size, and texture–gradient subsets by factors of $5\times$, $10\times$, and $5\times$, respectively. We further apply re-balanced angle augmentation to all upsampled data. Eventually, the effective number of ID-Constraint training samples per epoch is approximately 107,842. We visualize examples of both region-level supervision tasks and the proposed pixel-level supervision tasks in Figure~\ref{fig2:idcue}.

\subsection{Evaluation Benchmarks}
We evaluate our models on five standard MDE benchmarks: DIODE~\cite{vasiljevic2019diode}, ScanNet~\cite{dai2017scannet}, ETH3D~\cite{schops2017multi}, KITTI~\cite{geiger2012we}, and NYUv2~\cite{silberman2012indoor}. During implementation, we observed that existing state-of-the-art methods report results only on valid pixels. Following this convention, we first apply the provided validity masks of each dataset to filter out invalid depth regions, and then use the minimum and maximum depth thresholds to further filter and clamp depth values, consistent with the open-source GenPercept codebase~\cite{xu2025genpercept}. Besides, if an image contains fewer than 100 valid pixels, we treat it as corrupted and exclude it from evaluation. All rotations and resizing of depth maps and masks in preprocessing use nearest-neighbor interpolation to ensure valid-pixel integrity. The above dataset preprocessing might be different from DAv2~\cite{yang2024depthv2} and DistillAD~\cite{he2025distill}, as they don't release their evaluation code yet. The detailed dataset sizes and valid depth ranges for each benchmark are provided as follows:

\begin{itemize}
    \item \textbf{DIODE~\cite{vasiljevic2019diode}}: DIODE includes both indoor and outdoor scenes. After filtering with the validity masks, the test set contains 771 valid samples, with a valid depth range of $[0.6, 350]$.
    
    \item \textbf{ScanNet~\cite{dai2017scannet}}: ScanNet is a widely used indoor reconstruction dataset. After validity filtering, it contains 800 valid test samples, with a depth range of $[1\times 10^{-3}, 10]$.
    
    \item \textbf{ETH3D~\cite{schops2017multi}}: ETH3D provides high-resolution images with sparse depth annotations. After filtering, the test set contains 454 valid samples, with a valid depth range of $[1\times 10^{-5}, \infty)$.
    
    \item \textbf{KITTI~\cite{geiger2012we}}: KITTI is an outdoor dataset collected for autonomous driving, featuring images with a large field of view and therefore a very wide aspect ratio. After applying validity filtering, the test set contains 652 valid samples, with a valid depth range of $[1\times 10^{-5}, 80]$.
    
    \item \textbf{NYUv2~\cite{silberman2012indoor}}: NYUv2 consists of indoor scenes. After validity filtering, it includes 654 valid test samples, with a valid depth range of $[1\times 10^{-3}, 10]$.
\end{itemize}
All evaluation samples are downloaded from the data links provided by GenPercept~\cite{xu2025genpercept}.

\begin{figure*}[t]
    \centering
    \includegraphics[width=1.0\linewidth]{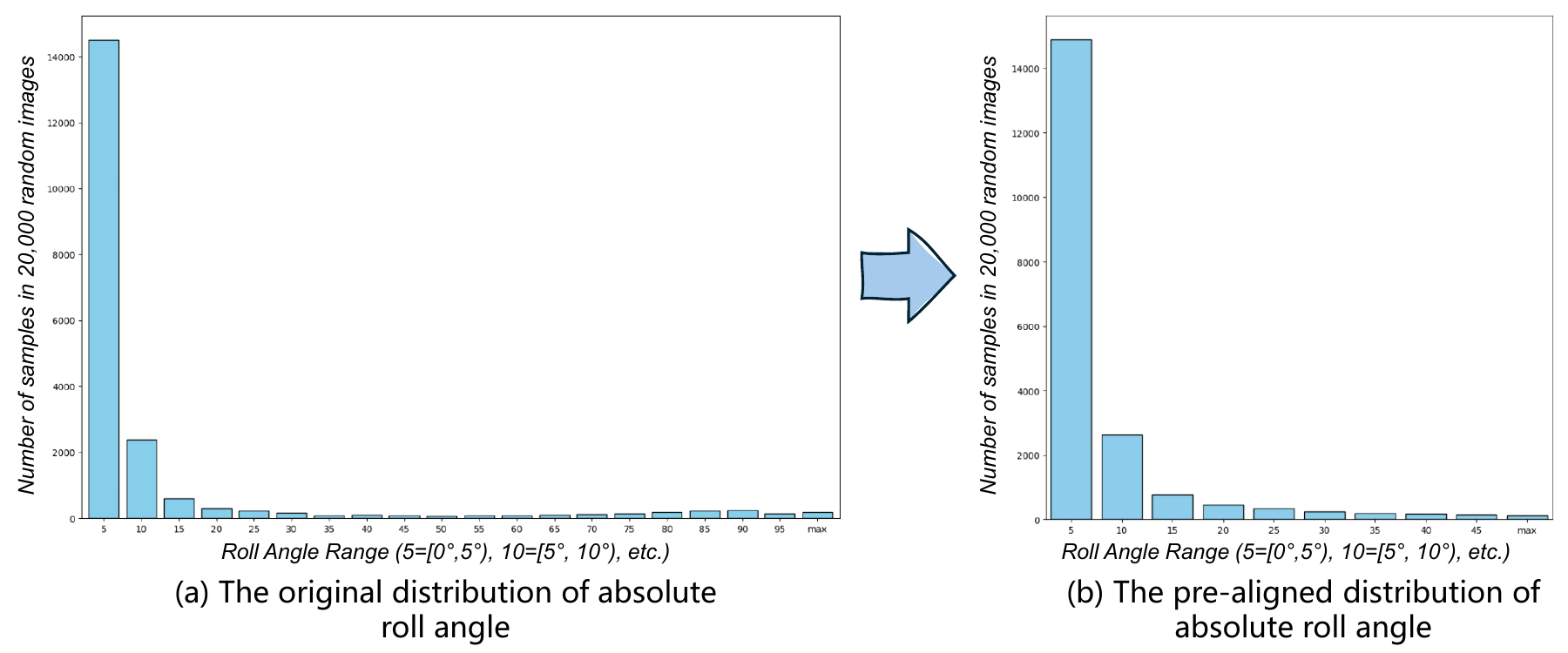}
    \caption{Distribution of absolute roll angles across 20,000 SA-1B samples. (a) The original distribution without any processing. (b) The pre-aligned distribution after applying a one-time $90^{\circ}$ rotation to samples whose detected roll angle exceeds $45^{\circ}$.}
    \label{figappx1:long-tail}
\end{figure*}

\section{Roll Angle Distribution Analysis}
\label{sec:appendixC}
To examine the distribution of absolute roll angles in real-world images, we apply our best roll-angle prediction model in Table 1 of the main paper to 20,000 SA-1B samples and visualize the resulting distribution in Figure~\ref{figappx1:long-tail}(a). During this process, we observed that a subset of images appears tipped due to incorrect post-processing, leading to a small suspicious peak near $90^{\circ}$. Since all of our angle prediction models reduce the absolute prediction error to below $48^{\circ}$, we can reliably correct these anomalies by applying a $90^{\circ}$ rotation to the affected samples. After this pre-alignment, the distribution is transformed into the one shown in Figure~\ref{figappx1:long-tail}(b), which we adopt as the true underlying training distribution used in the main paper.

Based on the above analysis, we can now better interpret the fine-grained roll-angle interval results shown in Figure 5 of the main paper and the detailed version in Figure~\ref{figappx2:intervals}.
Without horizon leveling, most methods exhibit a noticeable performance bump near $90^{\circ}$, which directly corresponds to the suspicious peak in the training distribution around the same angle.
In contrast, methods that incorporate horizon leveling, \textit{e.g.}, our proposed ID-Constraint approach, show an almost monotonic performance drop as the rolling angle increases. This pattern demonstrates that MDE performance is strongly correlated with the underlying data distribution, underscoring the severity of the horizontal prior bias in current MDE models.

\section{Additional Experimental Details}

\subsection{Experimental Environments}
All experiments are conducted in a Python 3.12, PyTorch 2.7, and TorchVision 0.22 environment, using a single A100-SXM4-80GB GPU. 

\subsection{Implementation Details of Figure 2 in the Main Paper} 
The motivation of this paper is because we observe severe degradation in our real-world commercial application using estimated depth to reconstruct the scene, where we found rolled frames cause edge blurs. Therefore, we found the horizontal prior problem. Due to the confidentiality agreement of the commercial project, I can only share part of the information here. We adopt the AGS-Mesh reconstruction~\cite{ren2025ags} that used a low-resolution sensor depth from mobile phone to recover the absolute scale and shift of the output. Since this setting is not normal MDE (more like a depth super-resolution task), therefore, we didn't add more results in the main paper.

\subsection{Implementation Details}
In this subsection, we are going to introduce the detailed implementation of the Invariant Depth Heads (ID Heads). ID Heads fall into two categories: region-level classification heads and pixel-level regression heads. The implementation of the classification heads follows \cite{danier2025depthcues}.
All heads take the multi-level feature maps $\{F_i\}$ extracted by $\text{ViT}(I)$ as input. The architectures of the ID Heads are detailed as follows:


\noindent\textbf{Light and shadow:} This task requires two object features to determine their belonging relationship. We first concatenate the multi-level inputs $\{F_i\}$ and pass them through a linear layer followed by bilinear upsampling to match the spatial resolution of the object masks $M_a$ and $M_b$, where $M_a, M_b \in \{0,1\}^{H \times W}$. We denote this process as $UP(\{F_i\})$. We then compute object-wise pooled features by masking and averaging over valid pixels. The final prediction is obtained as $Y_{\text{light-shadow}} = \text{MLP}(\frac{\sum^{H \times W} M_a \odot UP(\{F_i\})}{\sum^{H \times W} M_a} - \frac{\sum^{H \times W} M_b \odot UP(\{F_i\})}{\sum^{H \times W} M_b})$, where the MLP consists of two linear layers with a GELU activation between them, the output $Y_{\text{light-shadow}}$ is a binary classification score.

\noindent\textbf{Occlusion:} This task takes a single object as input and predicts whether it is occluded. Similar to the previous head, we apply an upsampling module $UP(\{F_i\})$ to obtain the upsampled feature map. The final prediction is computed as $Y_{\text{occlusion}} = \text{MLP}(\frac{\sum^{H \times W} M_a \odot UP(\{F_i\})}{\sum^{H \times W} M_a})$, where the MLP consists of two linear layers with a GELU activation, and $Y_{\text{occlusion}}$ is the binary classification output.

\noindent\textbf{Size:} This task requires two object features to determine their relative size, \textit{i.e.}, which object is larger. The decoder head shares the same structure as the light–shadow head, but with independently learned parameters. The final output $Y_{\text{size}}$ is a binary classification score.

\begin{table*}[t]
\centering
\fontsize{8}{10}\selectfont
\begin{tabular}{lccccccccccccccc} 
\hline
\toprule
\multirow{2}{*}{\textbf{Method}} 
          & \multicolumn{2}{c}{DIODE}
          &  & \multicolumn{2}{c}{ScanNet}
          &  & \multicolumn{2}{c}{ETH3d}
          &  & \multicolumn{2}{c}{KITTI}
          &  & \multicolumn{2}{c}{NYUv2}
          \\ 
    \cmidrule{2-3}\cmidrule{5-6}\cmidrule{8-9}\cmidrule{11-12} \cmidrule{14-15}

   & AbsRel$\downarrow$ & $\delta_1\uparrow$ & & AbsRel$\downarrow$ & $\delta_1\uparrow$ &
   & AbsRel$\downarrow$ & $\delta_1\uparrow$ & & AbsRel$\downarrow$ & $\delta_1\uparrow$ & 
   & AbsRel$\downarrow$ & $\delta_1\uparrow$ \\

\midrule

Marigold\textsuperscript{*}~\cite{ke2023marigold} & 0.279 & \textbf{0.776} & & 0.072 & 0.946 & & 0.068 & 0.956 & & 0.137 & 0.821 & & 0.061 & 0.958 \\
GenPercept\textsuperscript{*}~\cite{xu2025genpercept} & 0.297 & \underline{0.764} & & 0.062 & 0.959 & & 0.073 & 0.951 & & 0.130 & 0.841 & & 0.058 & 0.963 \\

\midrule

DAv2\textsuperscript{*}~\cite{yang2024depthv2} & \textbf{0.245} & 0.762 & & \textbf{0.042} & \textbf{0.979} & & \textbf{0.043} & \textbf{0.983} & & \underline{0.077} & \textbf{0.944} & & \textbf{0.044} & \textbf{0.979} \\
DistillAD\textsuperscript{*}~\cite{he2025distill} & \underline{0.246} & 0.758 & & 0.046 & \underline{0.978} & & 0.050 & 0.977 & & \textbf{0.074} & \underline{0.943} & & \underline{0.047} & \underline{0.977} \\

\midrule

(ours) Baseline & 0.252 & 0.756 & & 0.046 & 0.977 & & \underline{0.045} & \underline{0.981} & & \underline{0.077} & \underline{0.943} & & \underline{0.047} & \underline{0.977} \\

(ours) Re-balanced Aug & 0.251 & 0.758 & & 0.049 & 0.977 & & 0.058 & 0.973 & & 0.083 & 0.937 & & 0.051 & \underline{0.977} \\

(ours) Horizon Leveling  & 0.252 & 0.756 & & 0.046 & 0.977 & & \underline{0.045} & \underline{0.981} & & \underline{0.077} & \underline{0.943} & & \underline{0.047} & \underline{0.977} \\


(ours) ID-Constraint & 0.251 & 0.757 &  & \underline{0.045} & 0.977 &  & \textbf{0.043} & \underline{0.981} &  & \underline{0.077} & \textbf{0.944} &  & \underline{0.047} & \underline{0.977} \\

\bottomrule
\hline
\end{tabular}
\caption{Experimental results of state-of-the-art MDE models and the proposed algorithms under the \textbf{Horizontal ($0^{\circ}$)} setting. Superscript \textsuperscript{*} indicates results evaluated using our reimplemented evaluation code.}
\label{tab_appx1:horizon}
\end{table*}
\begin{table*}[t]
\centering
\fontsize{8}{10}\selectfont
\begin{tabular}{lccccccccccccccc} 
\hline
\toprule
\multirow{2}{*}{\textbf{Method}} 
          & \multicolumn{2}{c}{DIODE}
          &  & \multicolumn{2}{c}{ScanNet}
          &  & \multicolumn{2}{c}{ETH3d}
          &  & \multicolumn{2}{c}{KITTI}
          &  & \multicolumn{2}{c}{NYUv2}
          \\ 
    \cmidrule{2-3}\cmidrule{5-6}\cmidrule{8-9}\cmidrule{11-12} \cmidrule{14-15}

   & AbsRel$\downarrow$ & $\delta_1\uparrow$ & & AbsRel$\downarrow$ & $\delta_1\uparrow$ &
   & AbsRel$\downarrow$ & $\delta_1\uparrow$ & & AbsRel$\downarrow$ & $\delta_1\uparrow$ & 
   & AbsRel$\downarrow$ & $\delta_1\uparrow$ \\

\midrule

Marigold\textsuperscript{*}~\cite{ke2023marigold} & 0.294 & \textbf{0.758} & & 0.089 & 0.917 & & 0.080 & 0.939 & & 0.161 & 0.758 & & 0.069 & 0.950 \\
GenPercept\textsuperscript{*}~\cite{xu2025genpercept} & 0.306 & 0.752 & & 0.081 & 0.929 & & 0.080 & 0.944 & & 0.149 & 0.793 & & 0.064 & 0.956 \\

\midrule

DAv2\textsuperscript{*}~\cite{yang2024depthv2} & \textbf{0.250} & 0.755 & & \underline{0.049} & \underline{0.972} & & \underline{0.051} & \textbf{0.976} & & \underline{0.115} & 0.881 & & \textbf{0.049} & \textbf{0.977} \\
DistillAD\textsuperscript{*}~\cite{he2025distill} & \underline{0.252} & 0.750 & & 0.053 & 0.970 & & 0.059 & 0.969 & & 0.122 & 0.871 & & 0.053 & 0.974 \\

\midrule

(ours) Baseline & 0.256 & 0.751 & & 0.053 & 0.968 & & 0.055 & 0.971 & & 0.118 & 0.874 & & \underline{0.051} & 0.975 \\
(ours) Re-balanced Aug & 0.253 & \underline{0.757} & & \textbf{0.047} & \textbf{0.977} & & 0.056 & 0.973 & & 0.119 & 0.866 & & 0.052 & \underline{0.976} \\
(ours) Horizon Leveling  & 0.259 & 0.748 & & 0.059 & 0.959 & & 0.056 & 0.969 & & \textbf{0.080} & \underline{0.939} & & 0.053 & 0.970 \\
(ours) ID-Constraint & 0.254 & 0.754 &  & \underline{0.049} & \underline{0.972} &  & \textbf{0.049} & \underline{0.975} &  & \textbf{0.080} & \textbf{0.941} &  & \textbf{0.049} & 0.974 \\

\bottomrule
\hline
\end{tabular}
\caption{Experimental results of state-of-the-art MDE models and the proposed algorithms under the \textbf{Shaking [$0^{\circ},15^{\circ}$]} setting. Superscript \textsuperscript{*} indicates results evaluated using our reimplemented evaluation code.}
\label{tab_appx2:shaking}
\end{table*}
\begin{table*}[t]
\centering
\fontsize{8}{10}\selectfont
\begin{tabular}{lccccccccccccccc} 
\hline
\toprule
\multirow{2}{*}{\textbf{Method}} 
          & \multicolumn{2}{c}{DIODE}
          &  & \multicolumn{2}{c}{ScanNet}
          &  & \multicolumn{2}{c}{ETH3d}
          &  & \multicolumn{2}{c}{KITTI}
          &  & \multicolumn{2}{c}{NYUv2}
          \\ 
    \cmidrule{2-3}\cmidrule{5-6}\cmidrule{8-9}\cmidrule{11-12} \cmidrule{14-15}

   & AbsRel$\downarrow$ & $\delta_1\uparrow$ & & AbsRel$\downarrow$ & $\delta_1\uparrow$ &
   & AbsRel$\downarrow$ & $\delta_1\uparrow$ & & AbsRel$\downarrow$ & $\delta_1\uparrow$ & 
   & AbsRel$\downarrow$ & $\delta_1\uparrow$ \\

\midrule

Marigold\textsuperscript{*}~\cite{ke2023marigold} & 0.319 & 0.729 & & 0.110 & 0.878 & & 0.108 & 0.889 & & 0.198 & 0.682 & & 0.091 & 0.918 \\
GenPercept\textsuperscript{*}~\cite{xu2025genpercept} & 0.326 & 0.725 & & 0.094 & 0.908 & & 0.095 & 0.915 & & 0.190 & 0.697 & & 0.081 & 0.933 \\

\midrule

DAv2\textsuperscript{*}~\cite{yang2024depthv2} & 0.257 & 0.748 & & 0.064 & 0.956 & & 0.065 & 0.958 & & 0.118 & 0.877 & & 0.060 & 0.969 \\
DistillAD\textsuperscript{*}~\cite{he2025distill} & 0.261 & 0.744 & & 0.068 & 0.953 & & 0.073 & 0.951 & & 0.125 & 0.866 & & 0.066 & 0.963 \\

\midrule

(ours) Baseline & 0.267 & 0.742 & & 0.071 & 0.945 & & 0.073 & 0.948 & & 0.121 & 0.869 & & 0.064 & 0.964 \\
(ours) Re-balanced Aug & \underline{0.258} & \underline{0.753} & & \textbf{0.051} & \textbf{0.973} & & 0.062 & 0.964 & & \underline{0.115} & 0.875 & & 0.057 & \underline{0.972} \\
(ours) Horizon Leveling & 0.262 & 0.745 & & \underline{0.060} & 0.957 & & \underline{0.056} & \underline{0.968} & & \textbf{0.084} & \underline{0.934} & & \underline{0.056} & 0.966 \\
(ours) ID-Constraint & \textbf{0.255} & \textbf{0.754} &  & \textbf{0.051} & \underline{0.971} &  & \textbf{0.050} & \textbf{0.973} &  & \textbf{0.084} & \textbf{0.935} &  & \textbf{0.051} & \textbf{0.973} \\

\bottomrule
\hline
\end{tabular}
\caption{Experimental results of state-of-the-art MDE models and the proposed algorithms under the \textbf{Rolling [$0^{\circ},45^{\circ}$]} setting. Superscript \textsuperscript{*} indicates results evaluated using our reimplemented evaluation code.}
\label{tab_appx3:rolling}
\end{table*}

\begin{figure*}[t]
    \centering
    \includegraphics[width=1.0\linewidth]{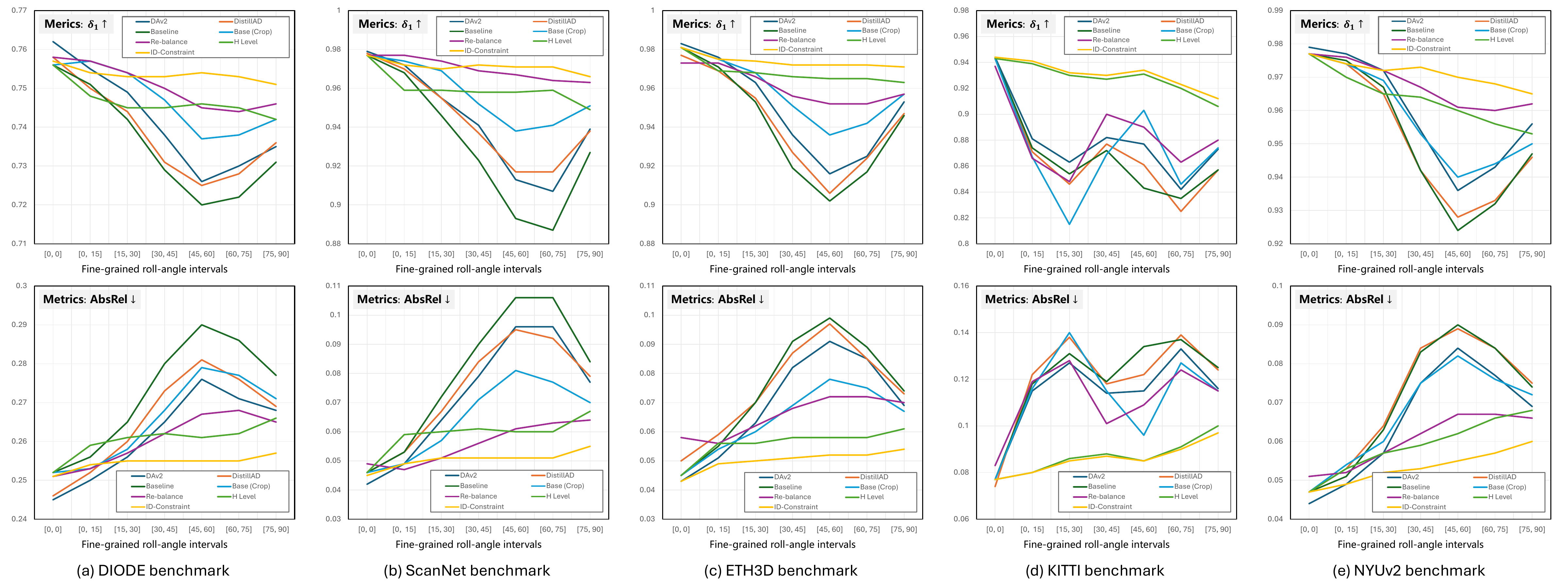}
    \caption{Experiments on fine-grained roll angle intervals for each benchmark. H Level stands for horizon leveling.}
    \label{figappx2:intervals}
\end{figure*}

\noindent\textbf{Texture gradient:} This task takes two pointer regions as input and predicts their relative depth, \textit{i.e.}, which point lies farther from the camera. The decoder head follows the same architecture as the light–shadow and size heads, but with separately learned parameters. The final output $Y_{\text{texture-grad}}$ is the binary classification output.

\noindent\textbf{Local peak and local slope:} For the local peak and local slope tasks, the ground-truth supervision consists of pixel-level peak maps and slope maps. Consequently, we adopt a DPT head~\cite{ranftl2021vision} as the task decoder, similar to the depth-prediction head. To reduce computational overhead, we set the hidden feature dimension of the DPT head to 128. We use the same decoder head to generate both outputs: $Y_{\text{local-ps}} = \text{DPT}(\{F_i\})$, where $Y_{\text{local-ps}} \in \mathbb{R}^{5\times H \times W}$. Among the five channels, one corresponds to the local peak prediction, and the remaining four correspond to local slope predictions along different directions. The final local-slope value is computed as the absolute average of the four directional slope outputs.

\subsection{Experimental Results}
In the main paper, due to space constraints, we report detailed results for the five benchmarks only under the most challenging Tipping [$0^{\circ},90^{\circ}$] setting. In this subsection, we provide the full benchmark results for other remaining three settings: Horizontal ($0^{\circ}$), Shaking [$0^{\circ},15^{\circ}$], and Rolling [$0^{\circ},45^{\circ}$]. As shown in Table~\ref{tab_appx1:horizon}, \ref{tab_appx2:shaking}, and \ref{tab_appx3:rolling},
DAv2~\cite{yang2024depthv2}, DistillAD~\cite{he2025distill} and Marigold~\cite{ke2023marigold} sometimes achieve stronger performance under the Horizontal and parts of the Shaking setting. Note that Horizontal is the conventional setting commonly adopted in prior work. Since their training code is not publicly released, we are unable to reproduce these results under our reimplemented training setup. In our own codebase, our proposed ID-Constraint approaches consistently outperform or remain competitive with our re-implemented baseline across all settings. This demonstrates that the proposed methods do not degrade performance in the Horizontal or Shaking settings when trained under the same conditions, confirming their robustness across a wide range of rolling angles.

To analyze how different roll angles affect model performance, Figure~\ref{figappx2:intervals} reports results on all five datasets across seven fine-grained angle intervals: [$0^{\circ},0^{\circ}$], [$0^{\circ},15^{\circ}$], [$15^{\circ},30^{\circ}$], [$30^{\circ},45^{\circ}$], [$45^{\circ},60^{\circ}$], [$60^{\circ},75^{\circ}$], and [$75^{\circ},90^{\circ}$]. A notable observation is a performance bump near $90^{\circ}$. This effect arises from several interacting factors: (1) \textbf{Training distribution.} Without horizon leveling, the training data itself exhibits a local distribution peak near $90^{\circ}$, which naturally leads to the performance increase in surrounding angles. A more in-depth analysis of the underlying causes of this local distribution peak is presented in Section~\ref{sec:appendixC}. (2) \textbf{Resolution change due to rotation.} Rotating an image from $0^{\circ}$ to $45^{\circ}$ increases its spatial resolution. Under a fixed number of patches in the ViT backbone, larger images compress more pixels into each patch, making depth prediction inherently more difficult. To isolate this effect, we introduce a Base (Crop) setting, where each rotated image is center-cropped to retain the same pixel numbers as the original image. Although Base (Crop) performs slightly better than the baseline, it follows the same overall trend, confirming that the horizontal prior is not an artifact of resolution changes. In addition, among all five benchmarks in Figure~\ref{figappx2:intervals}, a special case arises with the KITTI. As an outdoor autonomous-driving dataset, KITTI images have a very wide aspect ratio and large field of view, leading to more severe resolution distortion during rotation. Even Base (Crop) cannot fully compensate because aggressive cropping removes too much valid content. Consequently, as shown in Figure~\ref{figappx2:intervals}(d), only methods that explicitly apply horizon leveling maintain stable performance trends, making KITTI the most challenging benchmark in our experiments.

\begin{figure*}[t]
    \centering
    \includegraphics[width=1.0\linewidth]{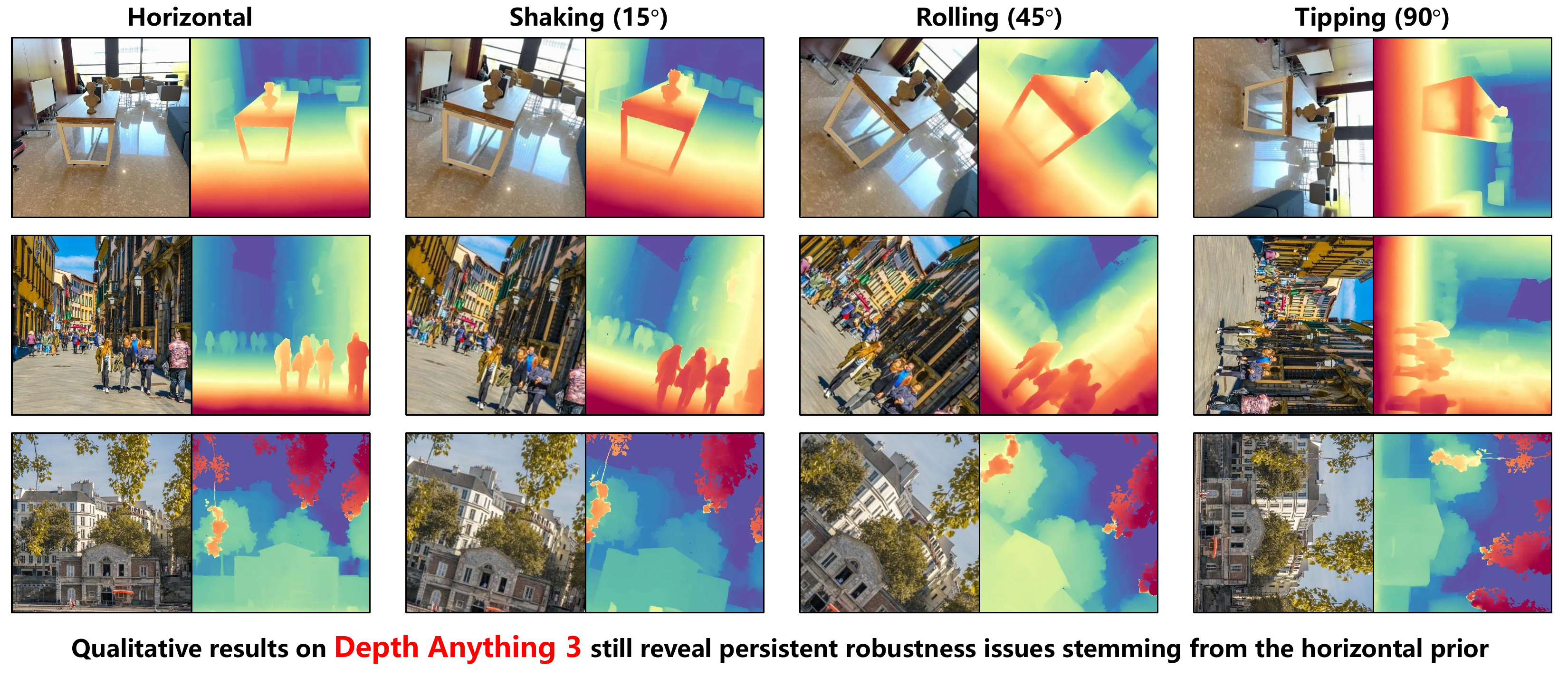}
    \caption{Qualitative analysis of the recently released Depth Anything 3, the current state-of-the-art model in MDE. Depth maps are generated from its huggingface online demo. The fuzzy boundaries observed under the Rolling and Tipping settings further confirm the persistent robustness issues induced by the horizontal prior.}
    \label{figappx8:dav3}
\end{figure*}

\begin{figure*}[t]
    \centering
    \includegraphics[width=1.0\linewidth]{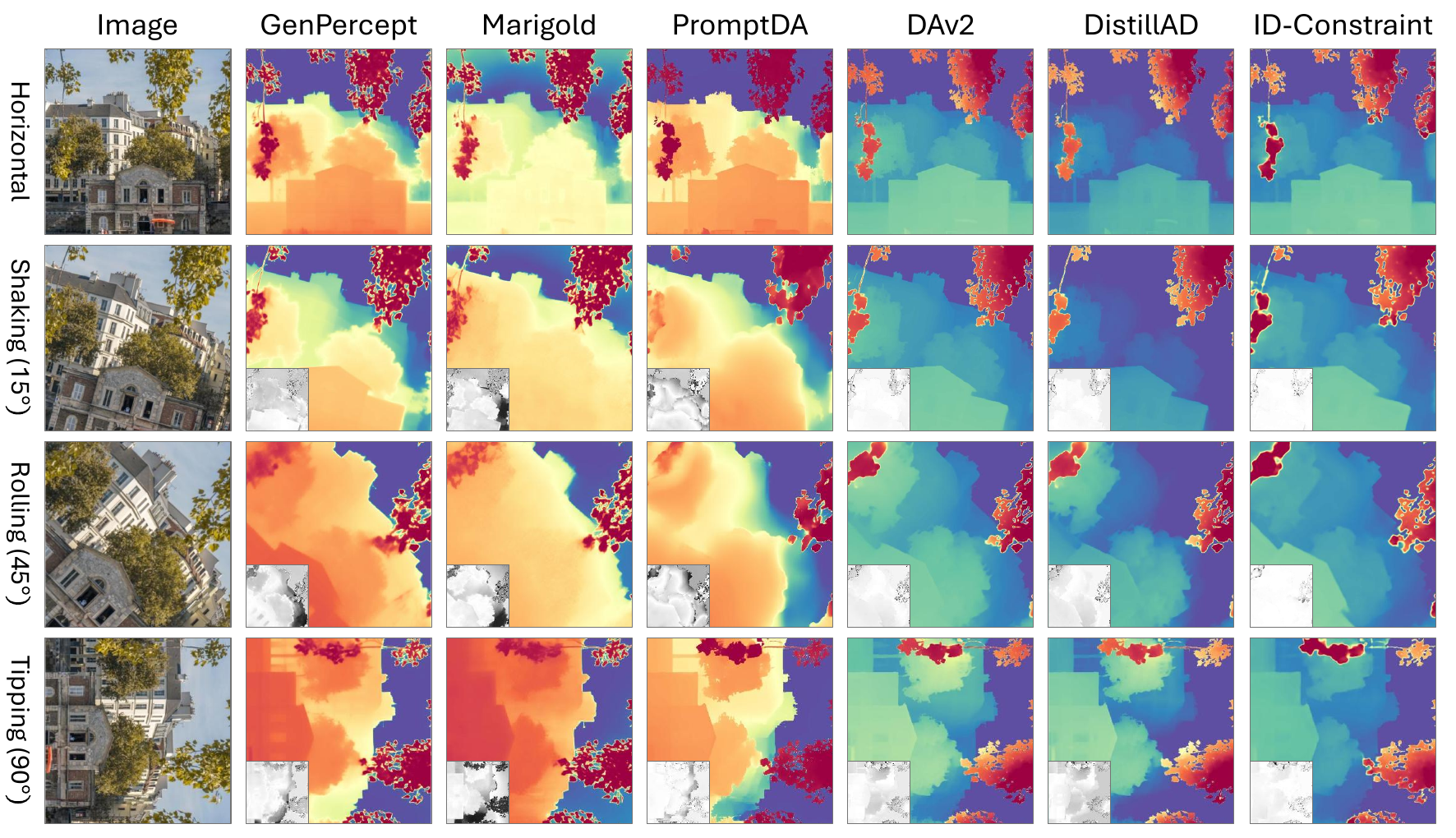}
    \caption{Qualitative comparisons of existing state-of-the-art MDE models and the proposed algorithms. The bottom-left grayscale images visualize the absolute error maps corresponding to their respective horizontal depth predictions. GenPercept, Marigold, and PromptDA produce outputs in the depth space, while the remaining output in the disparity space.}
    \label{figappx3:visualization}
\end{figure*}

\begin{figure*}[t]
    \centering
    \includegraphics[width=1.0\linewidth]{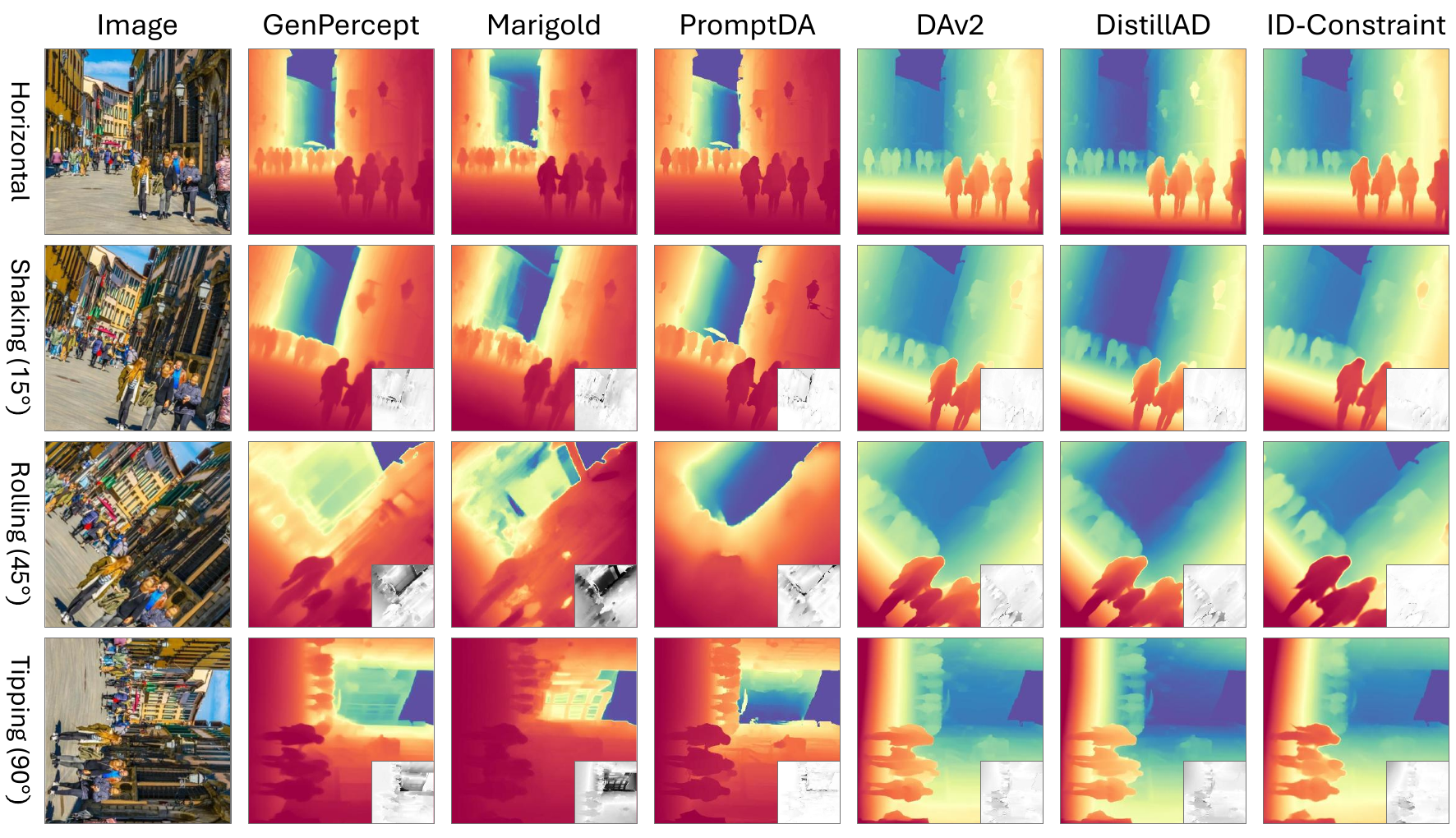}
    \caption{Qualitative comparisons of existing state-of-the-art MDE models and the proposed algorithms. The bottom-right grayscale images visualize the absolute error maps corresponding to their respective horizontal depth predictions. GenPercept, Marigold, and PromptDA produce outputs in the depth space, while the remaining output in the disparity space.}
    \label{figappx4:visualization}
\end{figure*}

\begin{figure*}[t]
    \centering
    \includegraphics[width=1.0\linewidth]{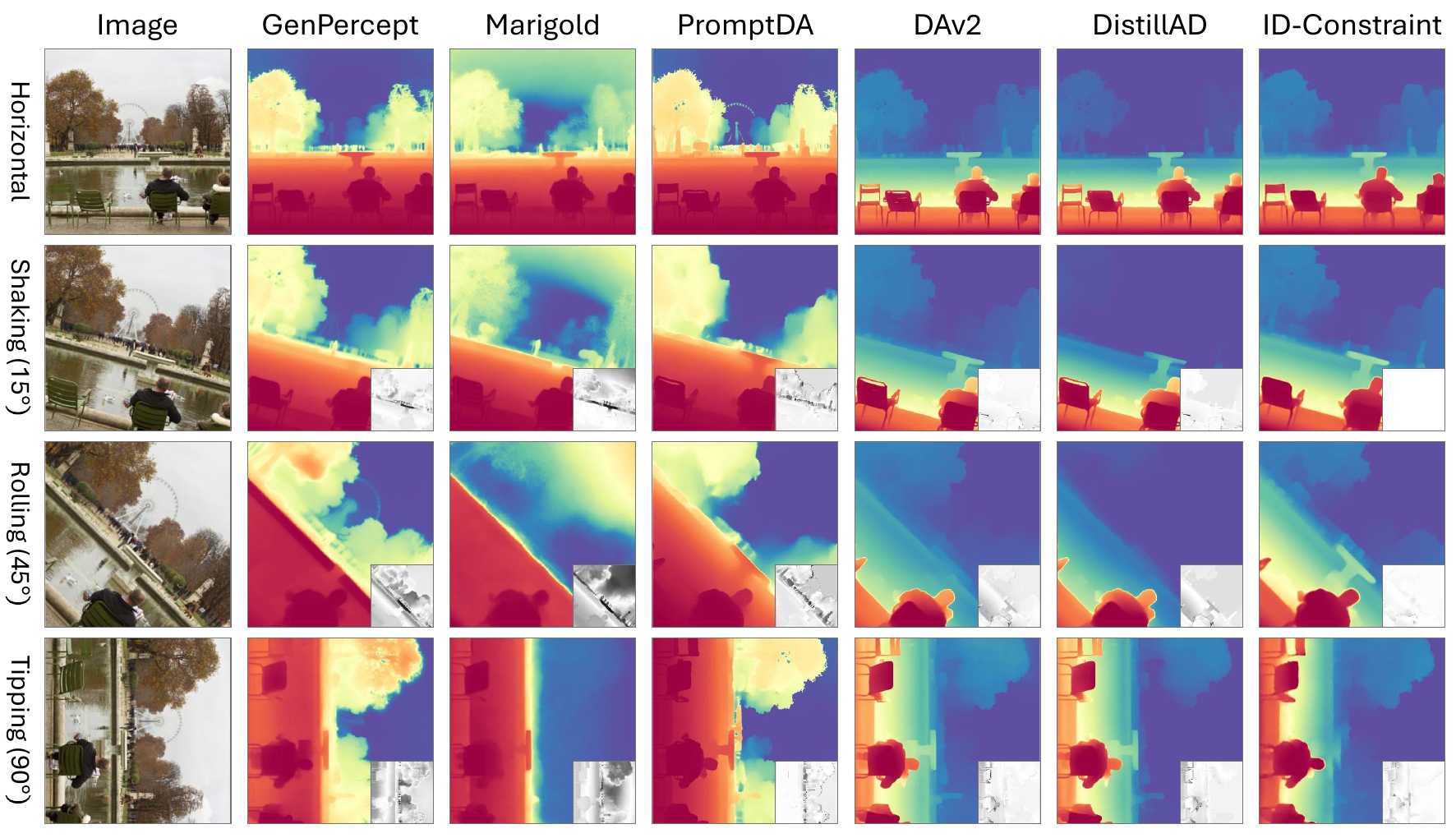}
    \caption{Qualitative comparisons of existing state-of-the-art MDE models and the proposed algorithms. The bottom-right grayscale images visualize the absolute error maps corresponding to their respective horizontal depth predictions. GenPercept, Marigold, and PromptDA produce outputs in the depth space, while the remaining output in the disparity space.}
    \label{figappx5:visualization}
\end{figure*}

\begin{figure*}[t]
    \centering
    \includegraphics[width=1.0\linewidth]{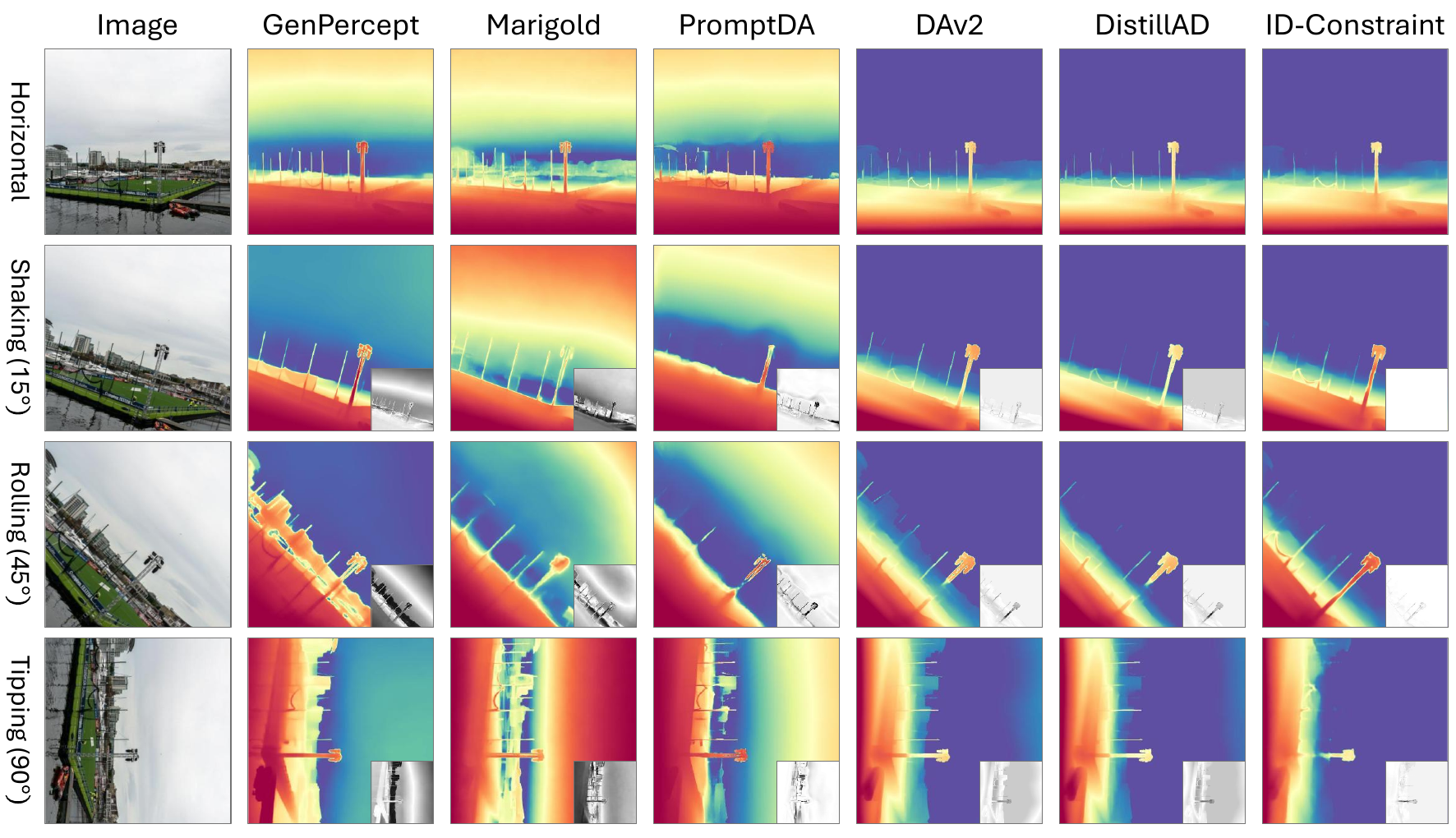}
    \caption{Qualitative comparisons of existing state-of-the-art MDE models and the proposed algorithms. The bottom-right grayscale images visualize the absolute error maps corresponding to their respective horizontal depth predictions. GenPercept, Marigold, and PromptDA produce outputs in the depth space, while the remaining output in the disparity space.}
    \label{figappx6:visualization}
\end{figure*}

\begin{figure*}[t]
    \centering
    \includegraphics[width=1.0\linewidth]{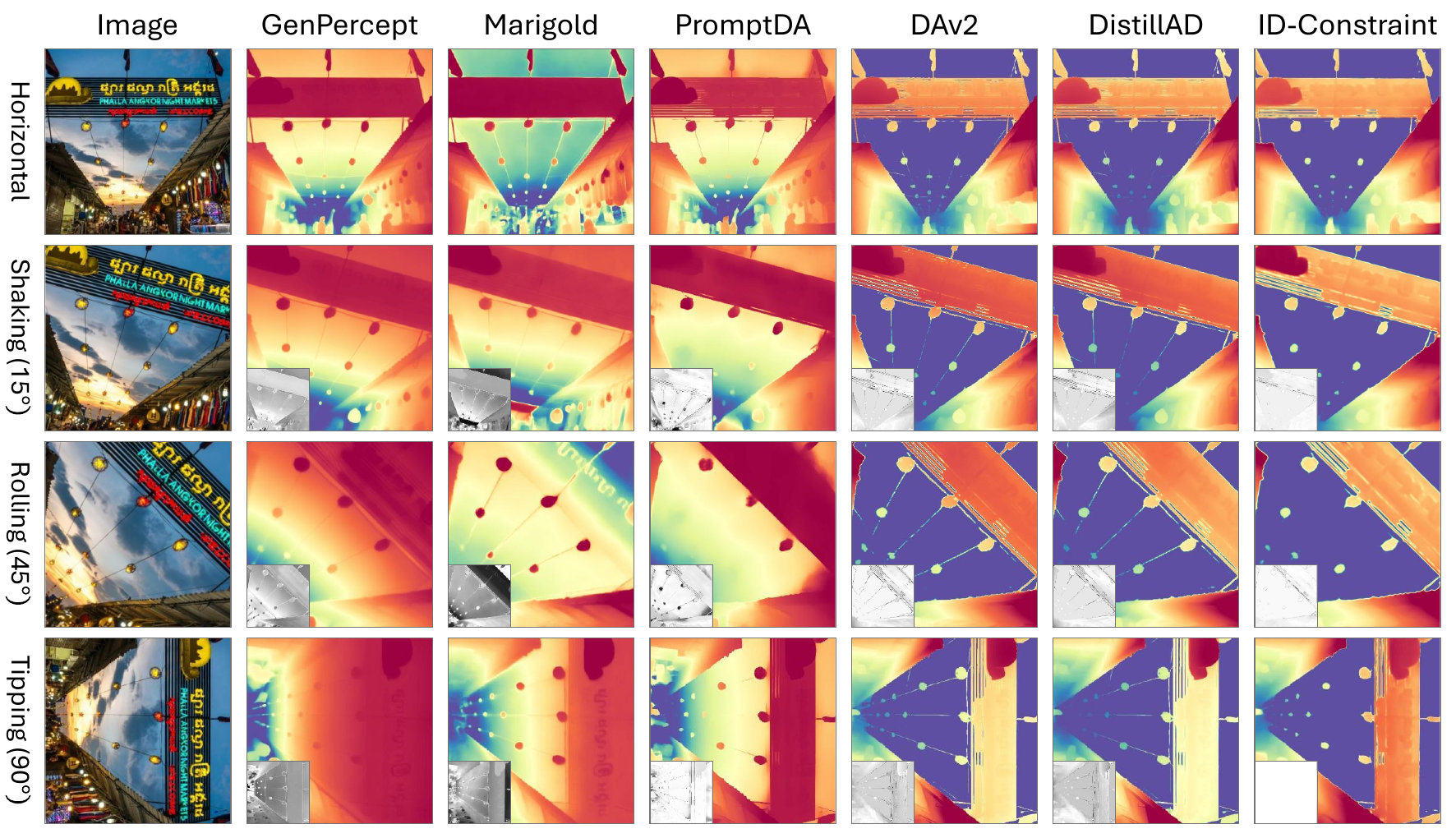}
    \caption{Qualitative comparisons of existing state-of-the-art MDE models and the proposed algorithms. The bottom-left grayscale images visualize the absolute error maps corresponding to their respective horizontal depth predictions. GenPercept, Marigold, and PromptDA produce outputs in the depth space, while the remaining output in the disparity space.}
    \label{figappx7:visualization}
\end{figure*}

\section{Discussion and Analysis}
\noindent\textbf{Definition of roll robustness.} We define roll robustness of a MDE model $f$ as $f(T_{\theta}(I)) \approx T_{\theta} (f(I)) $ ($\theta$-robust) under in-plane transform $T_{\theta>0^\circ}$. Degradation is measured by depth metrics. For Dataset averaged results: DAv2's AbsRel$\downarrow$ rises $0.096 \rightarrow 0.127$ (degrade $32\%$) and $\delta_1$$\uparrow$ falls $0.922 \rightarrow 0.883$ (degrade $3.9$ pts) between Horizontal and Tipping. For individual image tail case: (Figure 1(a) in the main paper, $\theta=45^\circ$): AbsRel$\downarrow$ degrades $45.7\%$ ($0.138 \rightarrow 0.201$), $\delta_1$$\uparrow$ degrades $8.7$ pts ($0.869 \rightarrow 0.782$) in scale and shift invariant domain, which are far worse than the average, confirming the severity.

\noindent\textbf{Motivation of choosing our two pixel-level tasks.} We chose local peak/slope because both are dense, local, affine-depth compatible, and rotation-invariant. Note that Surface normals also have potential but not a direct replacement here. A normal vector is equivariant (not invariant) under image roll because its components rotate with the image coordinate system. 

\noindent\textbf{Computational tradeoff analysis.} The proposed ID-Constraint adds zero inference parameters (auxiliary head only used during training). The only additional inference cost is horizon leveling, which adopts a ViT-S predictor ($\sim$21M) and +15.4 ms latency on one A100 with batch size 1. 

\noindent\textbf{Confidence level analysis.} We further added Paired Bootstrap Confidence Intervals to assess statistical reliability, where ID-Constraint \textit{vs.} Aug+level (95\% CI) has $\Delta \delta_1=+0.004 ~ [0.0026, 0.0052]$, $\Delta \text{AbsRel}=-0.004 ~ [-0.0047, -0.0031]$ on Tipping with 1K per-image bootstrap.

\noindent\textbf{Statistical grounding of long-tailed rolling bias.} The existence of horizontal prior rests on three independent pieces of evidence: 1) natural photographic data and averaged depth map in Figure 1 in the main paper; 2) multiple MDE models degrade when controlled roll increases; 3) the fine-grained roll analysis shows model behavior correlated with roll distribution. In addition, we also manually check those tail images and confirm they are genuinely tilted despite the $25.9^\circ$ error.  

\noindent\textbf{Stronger orientation estimation.} we adapted a recent method PerspectiveFields on our test data: mean roll error improves from  $25.9^\circ$ to $18.7^\circ$, yet, it is still far worse than the results reported in its original paper. We notice that they only evaluate roll estimation on GSV street dataset, which is a ``highly constrained setting'' (narrow roll range). Those cluttered indoor images in our dataset are much harder.

\section{Qualitative Visualizations}

\noindent\textbf{Qualitative results on Depth Anything 3.}
Recently, a new state-of-the-art MDE model, Depth Anything 3 (DA3), has been released, demonstrating strong performance over existing methods. Therefore, we provide a qualitative analysis generated by their online demo in this section. As shown in Figure~\ref{figappx8:dav3}, even this powerful state-of-the-art MDE model still exhibits robustness issues induced by the horizontal prior under the Rolling and Tipping settings, resulting in blurred boundaries around table legs and human shapes. This further demonstrates the importance of our study.

\noindent\textbf{Additional qualitative analyses.}
To further understand how different MDE models and the depth super-resolution model PromptDA~\cite{lin2025promptda} behave under various rolling conditions, we provide five additional visualization examples and notice several interesting observations: (1) PromptDA~\cite{lin2025promptda} is more susceptible to the Shaking and Rolling settings, even when provided with an additional low-resolution depth input. This is likely due to its training data being less diverse than that of DAv2~\cite{yang2024depthv2}, resulting in worse robustness. (2) More accurate roll-angle predictions lead to better ID-Constraint performance. For example, small prediction errors, \textit{e.g.}, below $2^{\circ}$ in Figure~\ref{figappx7:visualization} under the Tipping setting, yield strong ID-Constraint performance, whereas large errors, \textit{e.g.}, over $45^{\circ}$ in Figure~\ref{figappx6:visualization}, noticeably degrade performance. (3) ViT-based methods using the disparity space tend to be more robust than diffusion-based methods operating in depth space.

%
%
%


\end{document}